\documentclass{article}

\PassOptionsToPackage{numbers, compress}{natbib}
 \usepackage[preprint]{neurips_2026}

\usepackage[utf8]{inputenc} 
\usepackage[T1]{fontenc}    
\usepackage{hyperref}       
\usepackage{url}            
\usepackage{booktabs}       
\usepackage{amsmath}        
\usepackage{amsfonts}       
\usepackage{amssymb}       
\usepackage{amsthm}        
\usepackage[most]{tcolorbox} 
\usepackage{nicefrac}       
\usepackage{microtype}      
\usepackage[table]{xcolor} 
\usepackage[ruled,vlined]{algorithm2e}
\usepackage{tikz}
\usepackage{graphicx}
\usepackage{wrapfig}
\usepackage{capt-of}
\usepackage{float}
\usepackage{fancyhdr}
\AtBeginDocument{\DeclareMathSizes{10}{9}{7}{5}\DeclareMathSizes{9}{8}{6}{5}\DeclareMathSizes{12}{10}{7}{5}}
\newtcolorbox{proofbox}{
  colback=gray!8,
  colframe=gray!55,
  boxrule=0.5pt,
  arc=2pt,
  left=8pt,
  right=8pt,
  top=6pt,
  bottom=6pt,
  before skip=8pt,
  after skip=8pt,
  before upper={%
    \begingroup
    \AtBeginEnvironment{equation}{\footnotesize}
    \AtBeginEnvironment{align}{\small}
    \AtBeginEnvironment{gather}{\small}
    \AtBeginEnvironment{multline}{\small}},
  after upper={\endgroup},
  breakable
}

\definecolor{kvpoK}{RGB}{58,76,163}
\definecolor{kvpoV}{RGB}{66,99,170}
\definecolor{kvpoP}{RGB}{74,122,176}
\definecolor{kvpoO}{RGB}{82,145,183}
\definecolor{academicgreen}{RGB}{46,125,50}

\newcommand{\KVPOstyled}{\textbf{\textcolor{kvpoK}{K}\textcolor{kvpoV}{V}\textcolor{kvpoP}{P}\textcolor{kvpoO}{O}}}
\DeclareRobustCommand{\KVPO}{\texorpdfstring{\KVPOstyled}{KVPO}}
\newcommand{\upimprove}[1]{\textcolor{academicgreen}{($\Uparrow$#1\%)}}

\title{\KVPO{}: ODE-Native GRPO for Autoregressive Video Alignment via KV Semantic Exploration}

\author{%
  \begin{tabular}{c}
    {\large Ruicheng Zhang$^{1,3}$ \quad
    Kaixi Cong$^{1}$ \quad
    Jun Zhou$^{1}$ \quad
    Zhizhou Zhong$^{2,3}$} \\
    {\large Zunnan Xu$^{1}$ \quad
    Shuiyang Mao$^{3}$\thanks{Project leader.} \quad
    Wei Liu$^{3}$ \quad
    Xiu Li$^{1}$\thanks{Corresponding author.}}  \\[0.5em]
    {\normalsize $^{1}$Tsinghua University \quad
    $^{2}$HKUST \quad
    $^{3}$Video Rebirth} \\[0.5em]
  \end{tabular}%
}

\begin{document}

\setlength{\headheight}{32pt}
\thispagestyle{fancy}
\fancyhf{}
\vspace{-1em}
\fancyhead[L]{\raisebox{-0.35\height}{\includegraphics[height=37pt]{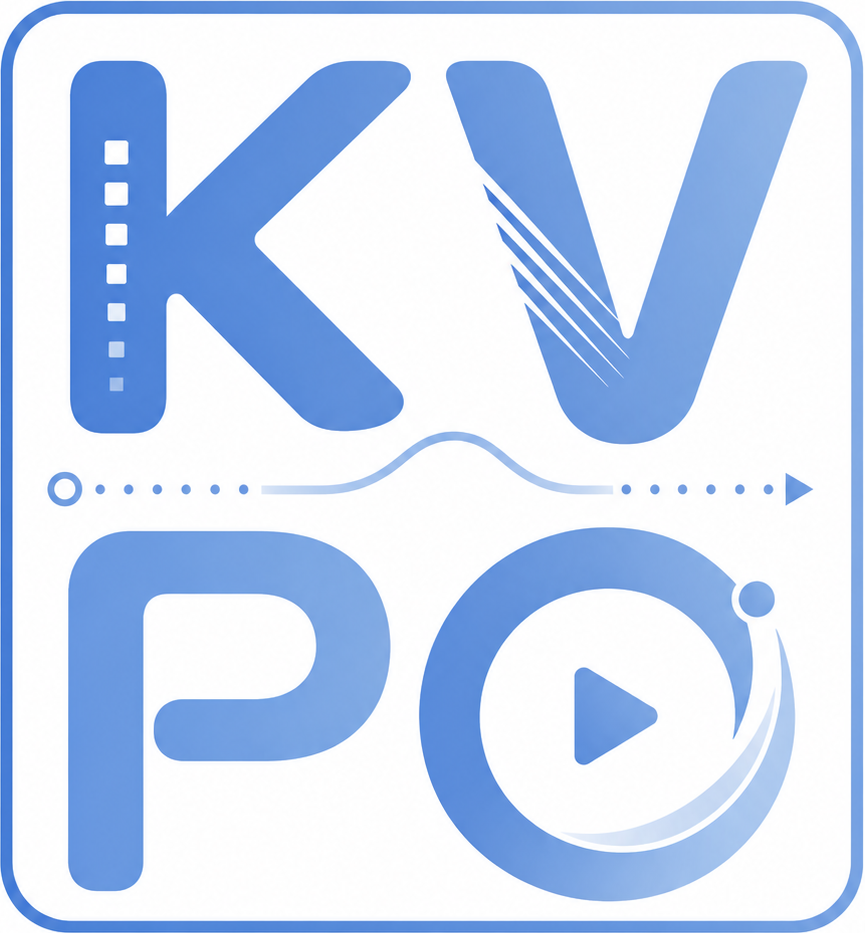}}}
\fancyhead[R]{\raisebox{-0.35\height}{\includegraphics[height=18pt]{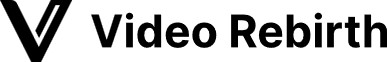}}}
\fancyfoot[C]{\thepage}
\renewcommand{\headrulewidth}{0pt}
\renewcommand{\headrule}{%
  \vskip 3.5pt%
  \hbox to\headwidth{\color{kvpoP}\leaders\hrule height 1.75pt\hfill}%
  \nointerlineskip
  \vskip 1.5pt%
  \hbox to\headwidth{\color{kvpoP}\leaders\hrule height 0.55pt\hfill}%
}

\begin{tcolorbox}[
  enhanced,
  colback=white,
  colframe=kvpoP,
  boxrule=1pt,
  arc=8pt,
  left=18pt,
  right=18pt,
  top=18pt,
  bottom=16pt,
  boxsep=0pt,
  width=\textwidth
]
{\raggedright
{\bfseries\LARGE \KVPO{}: ODE-Native GRPO for Autoregressive Video Alignment via KV Semantic Exploration\par}
\vspace{1.0em}

{\bfseries\normalsize
Ruicheng Zhang$^{1,3}$, Kaixi Cong$^{1}$, Jun Zhou$^{1}$, Zhizhou Zhong$^{2,3}$, Zunnan Xu$^{1}$,\par
 Shuiyang Mao$^{3}$\textsuperscript{\dag}, Wei Liu$^{3}$, Xiu Li$^{1}$\textsuperscript{\ddag}\par}
\vspace{0.65em}

{\bfseries\normalsize $^{1}$Tsinghua University, $^{2}$HKUST, $^{3}$Video Rebirth\par}
\vspace{0.35em}
{\small \textsuperscript{\dag}Project leader. \quad \textsuperscript{\ddag}Corresponding author.\par}
\vspace{1em}

{\normalsize
\begin{minipage}{\linewidth}
\sloppy
\setlength{\emergencystretch}{2em}
Aligning streaming autoregressive (AR) video generators with human preferences is challenging. Existing reinforcement learning methods predominantly rely on noise-based exploration and SDE-based surrogate policies that are mismatched to the deterministic ODE dynamics of distilled AR models, and tend to perturb low-level appearance rather than the high-level semantic storyline progression critical for long-horizon coherence. To address these limitations, we present \KVPO{}, an ODE-native online Group Relative Policy Optimization (GRPO) framework for aligning streaming video generators. For diversity exploration, \KVPO{} introduces a causal-semantic exploration paradigm that relocates the source of variation from stochastic noise to the historical KV cache. By stochastically routing historical KV entries, it constructs semantically diverse generation branches that remain strictly on the data manifold. For policy modeling, \KVPO{} introduces a velocity-field surrogate policy based on Trajectory Velocity Energy (TVE), which quantifies branch likelihood in flow-matching velocity space and yields a reward-weighted contrastive objective fully consistent with the native ODE formulation. Experiments on multiple distilled AR video generators demonstrate consistent gains in visual quality, motion quality, and text-video alignment across both single-prompt short-video and multi-prompt long-video settings.
\par
\end{minipage}}
\vspace{1em}

{\bfseries Homepage:} \url{https://richard-zhang-ai.github.io/KVPO-Project}\par
{\bfseries Code:} \url{https://github.com/Richard-Zhang-AI/KVPO}\par
{\bfseries Date:} \today\par
}
\end{tcolorbox}

\begin{center}
    \includegraphics[width=1\textwidth]{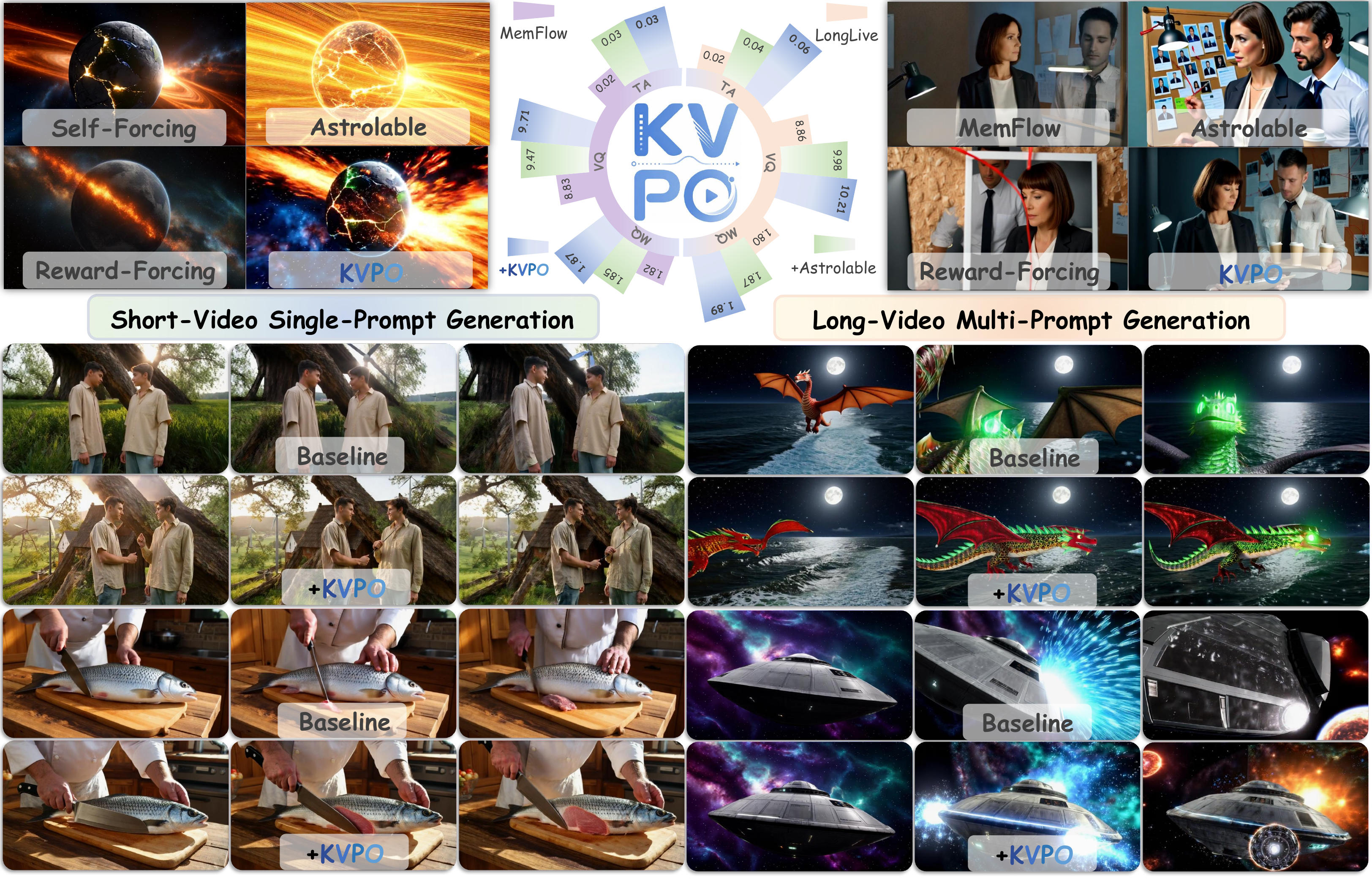}
    \vspace{-0.5em}
\end{center}

\section{Introduction}
\vspace{-0.5em}
\label{intro}

Recent advances in video generation~\cite{luo2025univid,zhang2026zo3t,hu2026identity,xu2025hunyuanportrait,zhang2026robostereo,zhang2025mind} have substantially improved visual quality, yet deploying these models in real-time interactive settings remains challenging. Such settings demand not merely high-fidelity generation, but low-latency, streaming, long-horizon synthesis under causal temporal dependencies. To meet these requirements, recent work distills pretrained video diffusion models into few-step autoregressive (AR) video generators, enabling efficient streaming inference via causal attention and KV caching~\cite{yin2024dmd,selfforcing,rewardforcing}. Nevertheless, aligning these AR video models with human preferences remains an open challenge, as preference-relevant qualities extend beyond frame-level fidelity to long-horizon coherence, subject consistency, and semantic progression.

\begin{wrapfigure}{r}{0.63\textwidth}
    \vspace{-1.5em}
    \centering
    \setlength{\abovecaptionskip}{-1.25em}
    \includegraphics[width=0.63\textwidth]{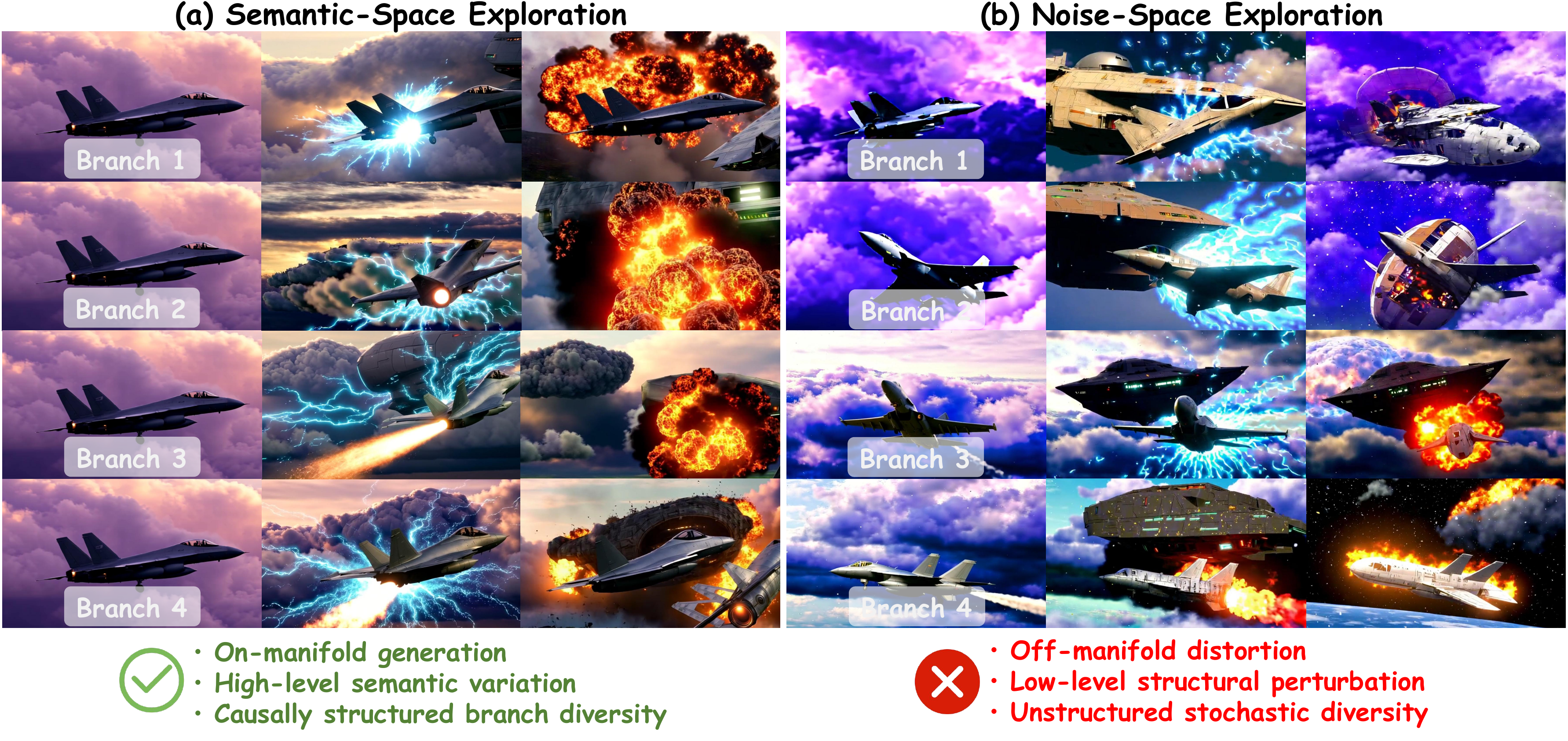}
    \caption{\textbf{Comparison of semantic-space and noise-space exploration.}}
    \label{fig:exploration}
    \vspace{-1.5em}
\end{wrapfigure}

Existing alignment methods for AR video generators predominantly fall into two categories, yet neither adequately addresses these challenges.
The first relies on reward-weighted distillation~\cite{rewardforcing}, which upweights high-reward trajectories in the supervised objective but fundamentally lacks active exploration of diverse candidate behaviors. 
The second~\cite{flowgrpo,dancegrpo} converts the deterministic ODE sampling into a stochastic SDE process and constructs exploration branches by injecting noise into the initial or intermediate latents. However, this strategy has been shown to be ill-suited to streaming AR video generators~\cite{he2025neighbor,he2026ar,sagegrpo}. Recasting a few-step distilled generator as an SDE injects stochastic transitions into an originally deterministic probability flow, which breaks its native ODE formulation~\cite{he2026ar}. Moreover, noise-driven exploration primarily perturbs low-level appearance and local structure~\cite{he2026ar} rather than the high-level semantics, motion dynamics, and storyline evolution that are crucial for long-horizon video generation (Figure~\ref{fig:exploration}). Furthermore, intermediate noise injection induces off-manifold structural interference~\cite{sagegrpo}, exacerbating the risk of generative degradation and weakening exploration signal quality.

More recently, NeighborGRPO~\cite{he2025neighbor} reinterprets Group Relative Policy Optimization (GRPO)~\cite{shao2024deepseekmath} as an implicit contrastive learning paradigm. It approximates the surrogate policy via Euclidean distances between samples generated under a pure ODE framework, with AR-CoPO~\cite{he2026ar} extending this approach to AR video generation. While this line of work offers useful insights into ODE-based policy optimization, surrogate policies grounded in latent Euclidean distances implicitly assume uniform geometry in the generation space, even though different latent dimensions may contribute unequally to policy probabilities. Therefore, such metrics may fail to faithfully capture the model's intrinsic preferences structure over candidate trajectories.

To overcome these limitations, we propose \KVPO{}, an ODE-native online GRPO framework tailored to streaming autoregressive video generation. \KVPO{} pioneers causal-semantic exploration and surrogate policy modeling in the flow-matching~\cite{lipman2023flow} velocity-field space under a pure ODE paradigm.
Unlike noise-driven perturbation approaches, we introduce a causal-semantic exploration paradigm that relocates the source of variation from stochastic noise to the historical KV cache. In streaming AR video generation, future content is causally conditioned on historical context, making differential reuse of historical information a natural mechanism for diversity exploration. Specifically, we design \textbf{Causal History Routing (CHR)}, which stochastically routes historical KV entries to construct branch-specific local contexts. Consequently, exploration remains strictly on-manifold, and variation in semantic space naturally promotes more meaningful and causally coherent narrative progression.
To optimize preferences over the explored branches, we further introduce an ODE-native surrogate policy formulation grounded in flow-matching dynamics. Rather than relying on external geometric distances or SDE transition kernels, we define a Gibbs-form surrogate policy based on \textbf{Trajectory Velocity Energy (TVE)} to quantify the likelihood of the current policy reproducing each branch directly in the velocity-field space.
This yields a reward-weighted contrastive flow-matching objective that embeds preference optimization into the model's native dynamics.

Experiments on multiple distilled AR video generators demonstrate consistent gains in human-preference alignment across both single-prompt short-video and multi-prompt long-video settings. Our primary contributions are as follows:
\vspace{-0.5em}
\begin{itemize}
    \item We propose \KVPO{}, an ODE-native online policy optimization framework for streaming AR video generation. To the best of our knowledge, \KVPO{} is the first method to perform causal-semantic exploration and model the surrogate policy within the flow-matching velocity-field space under a pure ODE paradigm.
    \item We introduce a causal-semantic exploration mechanism that shifts diversity generation from unstructured noise injection to historical KV-cache routing, intrinsically avoiding off-manifold distortion while promoting richer narrative progression and storyline diversity.
    \item We introduce a velocity-field surrogate policy based on Trajectory Velocity Energy (TVE), which yields a reward-weighted contrastive flow-matching objective that embeds preference optimization into the model's native ODE dynamics without relying on external geometric distances or SDE transition kernels.
\end{itemize}

\vspace{-0.5em}
\section{Related Work}
\vspace{-0.5em}

\subsection{Streaming Autoregressive Video Generation}
\vspace{-0.5em}
Autoregressive (AR) models~\cite{yin2024onestep,yin2025causvid} generate video in a causal, streaming fashion by conditioning each new frame on previously generated content. Recent acceleration and distillation techniques have substantially improved their practicality, compressing multi-step diffusion processes into efficient few-step variants while preserving visual quality~\cite{yin2024dmd,song2023consistency,zhao2023unipc}. By exploiting causal attention, dynamic key-value (KV) caching~\cite{ji2026forcingkvhybridkvcache}, and explicit memory architectures~\cite{memflow}, these models enable interactive, real-time, and long-horizon video generation~\cite{selfforcing,longlive,memflow}. Despite these advances, explicit preference alignment for highly deterministic few-step AR models remains relatively underexplored.

\vspace{-0.5em}
\subsection{Preference Alignment for Generative Models}
\vspace{-0.5em}
Post-training alignment for generative models typically leverages reward signals to steer model outputs toward human-preferred behaviors. This is commonly achieved by framing the sampling process as a policy rollout and optimizing the induced distribution via policy-gradient objectives.
VideoAlign~\cite{videoalign} introduces reward supervision for video generation. Flow-GRPO~\cite{flowgrpo} and DanceGRPO~\cite{dancegrpo} extend GRPO-style optimization to visual generative models by reformulating ODEs as SDEs. However, such noise-injection rollout strategies and SDE-based policy modeling paradigms are ill-suited for few-step AR video models~\cite{he2026ar}. These methods deviate from the native ODE formulation of AR generators and tend to alter low-level appearance more than high-level semantic development. SAGE-GRPO~\cite{sagegrpo} further shows that noise-based exploration can induce off-manifold distortions, undermining the quality of candidate samples.

Recent works have begun to explore alignment techniques tailored for AR video models. Reward Forcing~\cite{rewardforcing} performs reward-weighted distillation to amplify optimization signals from high-quality samples, but lacks active exploration. Astrolabe~\cite{astrolabe} applies forward-process reinforcement learning by contrasting positive and negative samples at inference endpoints, yet exploration remains confined to noise-endpoint perturbation rather than structured semantic branching. NeighborGRPO~\cite{he2025neighbor} offers an ODE-centric alternative by modeling preferences through latent-space neighborhood geometry, and AR-CoPO~\cite{he2026ar} extends this to AR video generation. Nevertheless, both depend on external geometric proximity to approximate surrogate preference ordering, which may not faithfully reflect the model's intrinsic preferences over candidate trajectories. In contrast, \KVPO{} performs causal-semantic exploration via stochastic KV routing and models the surrogate policy in the ODE-native flow-matching velocity-field space, offering a new perspective on AR preference alignment.

\vspace{-0.5em}
\section{Methodology}
\vspace{-0.5em}
\label{sec:method}

\subsection{Preliminaries: Block-wise Autoregressive Video Generation}
\vspace{-0.5em}
\label{subsec:preliminaries}

Mainstream streaming AR video generators synthesize long videos in a block-by-block manner. Given a video sequence $V = [v_1, v_2, \dots, v_B]$ partitioned into $B$ blocks, the generation at block $b$ is formulated as $p_\theta(v_b \mid v_{<b}, \mathcal{C})$, conditioned on the text prompt $\mathcal{C}$ and the historical context $v_{<b}$. In Diffusion Transformer (DiT)~\cite{dit} architectures, this historical context is materialized as a compressed Key-Value (KV) cache $\mathcal{K}_{<b}$. In streaming implementations, the KV memory typically adopts a $(\mathrm{sink},\, \mathrm{local})$ structure: the sink cache stores persistent global anchors for long-range temporal coherence, while the local cache maintains a sliding window of the $N$ most recent frames for local motion modeling.
Under the flow matching framework~\cite{lipman2023flow}, the model is trained along the linear interpolation path between a clean sample $x_0$ and a noise latent $x_T$:
\begin{equation}
    x_t = t\,x_0 + (1-t)\,x_T, \quad x_T \sim \mathcal{N}(0,\mathbf{I}),
    \quad t \in [0,1].
\end{equation}
A conditional velocity field $v_\theta(x_t, t, \mathcal{K}_{<b})$ is learned by minimizing the expected squared error against the ground-truth velocity $u_t = x_0 - x_T$. At inference, block $x^b_0$ is obtained by integrating the probability flow ODE from noise to clean:
\begin{equation}
    \frac{dx_t}{dt} = v_\theta(x_t,\, t,\, \mathcal{K}_{<b}),
    \quad x_{t=0} = x_T \sim \mathcal{N}(0, \mathbf{I}).
\end{equation}
The ODE solver advances through $S$ discrete timesteps $\{t_s\}_{s=1}^S$, yielding the generated block $x^b_0 = v_b$.

\subsection{Causal-Semantic Exploration via Causal History Routing}
\vspace{-0.5em}
\label{subsec:causal_routing}

We redirect diversity exploration from noise-driven perturbations to causal-semantic exploration over the historical KV cache via \textbf{Causal History Routing (CHR)}. Since future content in streaming AR video generation is strongly conditioned on the historical context $\mathcal{K}_{<b}$, perturbing the composition of local memory induces semantically diverse generation branches. Specifically, consider a pivot block $b^*$ at which $L$ frames have already been generated. CHR leaves the sink KV unchanged, where the sink memory comprises the earliest three historical frames: $\mathcal{K}_{\mathrm{sink}} = \{(K_1,V_1), (K_2,V_2), (K_3,V_3)\}$. For local memory, CHR adopts a fixed $9$-slot layout in which the last three slots always store the most recent frames, $\mathcal{K}_{\mathrm{near}} = \{(K_{L-2},V_{L-2}), (K_{L-1},V_{L-1}), (K_L,V_L)\}$, while the first six slots are branch-specific and stochastically refilled from the older non-sink history. Letting $\Omega_L = \{4,5,\dots,L-3\}$ denote the routable index set, CHR samples six indices $r_1^g,\dots,r_6^g \in \Omega_L$ for each branch $g\in\{1,\dots,G\}$ and constructs the branch-specific local cache as
\begin{equation}
    \tilde{\mathcal{K}}_{<b^*}^{g,\mathrm{local}}
    = \Bigl[
    \underbrace{(K_{r_1^g},V_{r_1^g}), \dots, (K_{r_6^g},V_{r_6^g})}_{\text{branch-specific routed 6 slots}}
    \,;\,
    \underbrace{\mathcal{K}_{\mathrm{near}}}_{\text{shared recent 3 slots}}
    \Bigr].
\end{equation}
For each candidate branch $g$, the attention output at block $b^*$ is computed using the current-block query $Q_{b^*}$ against the concatenation of the sink cache, the branch-specific local cache, and the current-block KV entries:
\begin{equation}
    \mathrm{Attn}^{g}_{b^*}
    = \mathrm{Softmax}\!\bigl(
        \frac{Q_{b^*}^{g}\,\bigl[ K_{\mathrm{sink}}\,;\, \tilde{K}_{<b^*}^{g,\mathrm{local}}\,;\, K_{b^*}^{g} \bigr]^\top}{\sqrt{d_k}}
    \bigr)
    \bigl[ V_{\mathrm{sink}}\,;\, \tilde{V}_{<b^*}^{g,\mathrm{local}}\,;\, V_{b^*}^{g} \bigr],
    \label{eq:kv_perturbation}
\end{equation}
where $d_k$ denotes the key dimension.

\textbf{Rollout and Replay.} During rollout, semantic exploration branches from a randomly sampled pivot block $b^*$ under $G$ distinct CHR refill decisions for the branch-specific local slots. Blocks preceding $b^*$ are generated once using the shared default KV cache, while CHR is applied exclusively within a contiguous window $\mathcal{B} = [b^*, b^* + W)$, where $W$ denotes the exploration window length in blocks. Beyond $\mathcal{B}$, generation reverts to the standard local cache, yet the semantic variations introduced within the window propagate through subsequent blocks, as the perturbed KV states are written back into the cache.
Within each perturbed block, CHR is restricted to the first half of the ODE steps, motivated by the observation that early-to-mid solver stages govern coarse semantic layout and motion, whereas late-stage perturbations contribute marginally to semantic diversity while incurring unnecessary replay cost~\cite{he2026ar}.
The rollout produces $G$ branch trajectories $\{X^{g}\}$ with associated rewards $\{r^{g}\}$, alongside an anchor trajectory $X^{0}$ generated under the default local cache without CHR routing, yielding a baseline reward $r^0$. For each branch $g$ and solver step $s$, we cache replay tuples $\{z_{b,s}^{g},\, \hat{u}_{b,s}^{g}\}$ over the perturbed window $\mathcal{B}$, where $z_{b,s}^{g}$ denotes the intermediate latent at block $b$ and step $s$, and $\hat{u}_{b,s}^{g}$ the corresponding rollout velocity target.

During replay, the cached intermediate states $z_{b,s}^{g}$ from each branch are reused as input under the restored unperturbed context $\mathcal{K}_{<b}$ to predict replayed velocities $v_\theta(z_{b,s}^{g}, t_s, \mathcal{K}_{<b})$, which are subsequently used for surrogate policy modeling (Section~\ref{subsec:policy_modeling}). This procedure assesses the current model's generative tendency toward each branch trajectory under the unperturbed deployment-time semantics. Each replay step incurs the computational cost of a single forward pass and requires no specialized solver, making the replay stage as efficient as standard supervised fine-tuning. Gradient tracking is enabled exclusively for solver steps within the perturbed window $b \in \mathcal{B}$.

\begin{figure*}[t]
    \centering
     \setlength{\abovecaptionskip}{-1.25em}   
    \includegraphics[width=\textwidth]{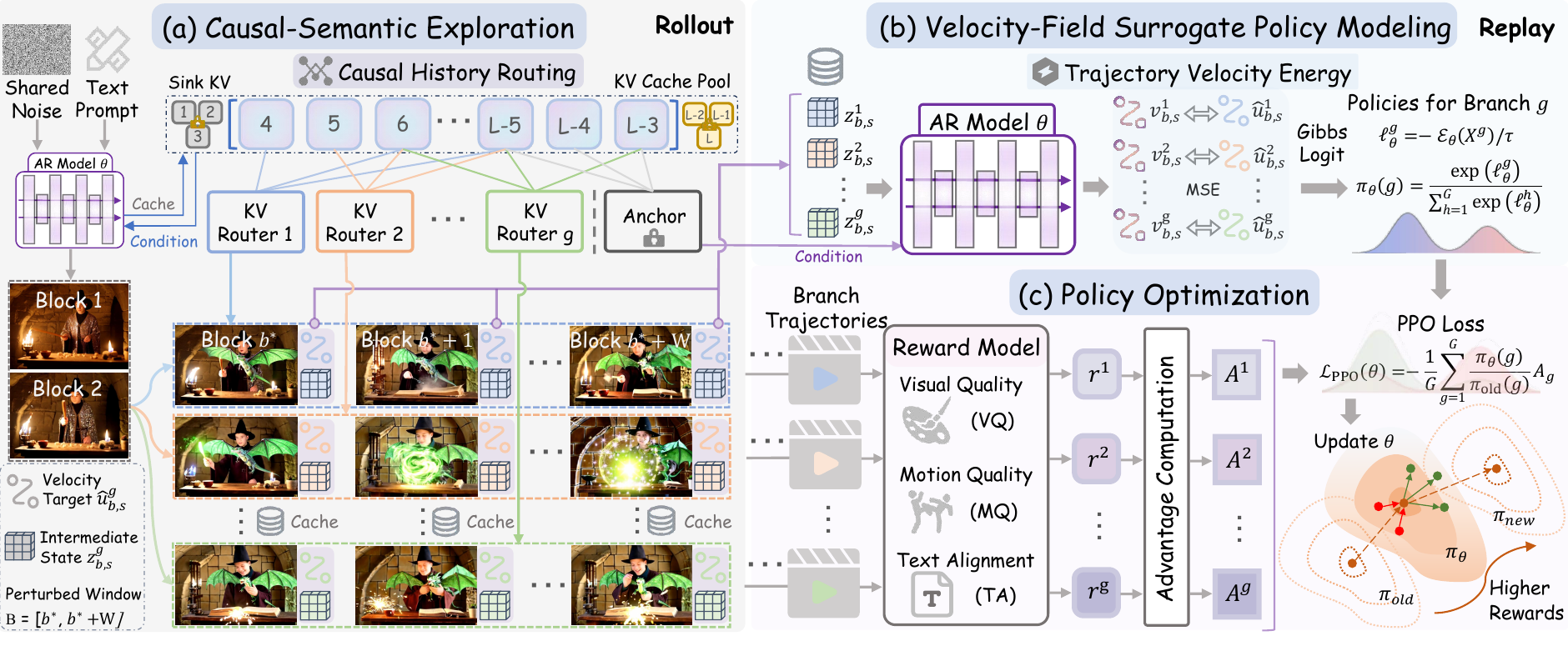}
    \caption{\textbf{Overview of the \KVPO{} training pipeline.} Starting from a shared initial noise, the model first performs causal-semantic exploration via stochastic KV routing within a perturbed window to produce diverse candidate branches~(a). These branches are then replayed under the unperturbed deployment-time context, where the Trajectory Velocity Energy of each branch is computed and converted into Gibbs-form surrogate branch probabilities to measure their generation likelihood under the current policy~(b). Finally, the branches are scored by the reward model, and PPO updates the AR generator toward higher-reward behaviors via a contrastive flow-matching objective~(c)}
    \label{fig:kvpo_framework}
    \vspace{-1em}
\end{figure*}

\subsection{Velocity-Field Surrogate Policy Modeling and Optimization}
\vspace{-0.5em}
\label{subsec:policy_modeling}

Deterministic ODE generators do not expose an explicit policy distribution over candidate branches, making direct application of PPO intractable~\cite{he2025neighbor}. Prior work~\cite{he2025neighbor,he2026ar} has shown that GRPO admits an interpretation as implicit contrastive learning: the update promotes reward-preferred generations while suppressing reward-disfavored ones via their relative advantages. Guided by this insight, we introduce a branch-wise quantity that captures the current model's generative likelihood under causal-semantic exploration and use it to construct a surrogate policy for preference optimization.

\textbf{Trajectory Velocity Energy (TVE).}
In \KVPO{}, causal-semantic exploration generates diverse candidate branches via stochastic local KV routing, while inference is performed under the unperturbed context $\mathcal{K}_{<b}$. The key quantity of interest is therefore the likelihood of the current policy reproducing the cached rollout velocities of a given branch under the unperturbed deployment-time semantics, which motivates the definition of Trajectory Velocity Energy.
Formally, TVE for branch trajectory $X^{g}$ is defined as the aggregated squared residual between the cached rollout velocity target $\hat{u}_{b,s}^{g}$ and the corresponding replayed velocity $v_\theta(z_{b,s}^{g},\, t_s,\, \mathcal{K}_{<b})$ across all perturbed blocks and solver steps:
\begin{equation}
    \mathcal{E}_\theta\!\left(X^{g}\right) 
    = \sum_{b \in \mathcal{B}} \sum_{s=1}^{S} 
    \frac{1}{d} 
    \left\| v_\theta\!\left(z_{b,s}^{g},\, t_s,\, 
    \mathcal{K}_{<b}\right) 
    - \hat{u}_{b,s}^{g} \right\|_F^2,
    \label{eq:tve}
\end{equation}
where $d$ denotes the feature dimension. TVE directly reflects branch likelihood in the flow-matching velocity space: a lower TVE indicates that the current policy assigns stronger generative tendency toward that branch under the unperturbed deployment-time context $\mathcal{K}_{<b}$.

\textbf{Surrogate Policy and Policy Ratio.} Having defined TVE as a measure of branch likelihood under the unperturbed deployment-time context, we convert these energy values into a normalized branch distribution to construct a surrogate policy. Such a conversion should satisfy three requirements: (1) branches with lower TVE receive higher policy probability; (2) the policy is differentiable and amenable to gradient optimization; and (3) the policy depends only on relative TVE scores across branches, aligning with the contrastive learning objective. Gibbs parameterization naturally satisfies all three. 
Let $\ell_\theta^{g} = -\mathcal{E}_\theta(X^{g}) / \tau$, where $\tau$ is 
a temperature parameter. The current and previous policies for branch $g$ are then defined as
\begin{equation}
    \pi_\theta(g)
    = \frac{\exp\!\left(\ell_\theta^{g}\right)}
           {\sum_{h=1}^{G}\exp\!\left(\ell_\theta^{h}\right)},
    \qquad
    \pi_{\mathrm{old}}(g)
    = \frac{\exp\!\left(\ell_{\mathrm{old}}^{g}\right)}
           {\sum_{h=1}^{G}\exp\!\left(\ell_{\mathrm{old}}^{h}\right)}.
    \label{eq:gibbs_policy}
\end{equation}
The resulting Gibbs distribution converts the model's generative tendencies into a normalized branch distribution. Unlike geometry-based surrogate policies~\cite{he2025neighbor}, our branch probabilities are grounded directly in replay-time compatibility, remaining faithful to the flow-matching model's native dynamics. The PPO importance ratio $\rho^{g} = \pi_\theta(g)/\pi_{\mathrm{old}}(g)$ is computed in the logarithmic domain as
\begin{equation}
    \log \rho^{g} = \log \pi_\theta(g) - \log \pi_{\mathrm{old}}(g) = (\ell_\theta^{g} - \log \sum_{h=1}^G \exp(\ell_\theta^{h})) - (\ell_{\mathrm{old}}^{g} - \log \sum_{h=1}^G \exp(\ell_{\mathrm{old}}^{h})).
    \label{eq:log_ratio}
\end{equation}
The generator parameters are then updated via the clipped PPO objective
\begin{equation}
    \mathcal{L}_{\text{PPO}}(\theta) = -\frac{1}{G} \sum_{g=1}^G \min \left( \rho^{g} A^{g}, \mathrm{clip}(\rho^{g}, 1-\epsilon_{\mathrm{low}}, 1+\epsilon_{\mathrm{high}}) A^{g} \right),
    \label{eq:ppo_objective}
\end{equation}
where the normalized branch advantage is
\begin{equation}
    A^{g} = \frac{r^{g} - \bar{r}}{\sqrt{\frac{1}{G} \sum_{k=1}^{G} \left(r^{k} - \bar{r}\right)^2} + \epsilon}, \quad \text{where } \bar{r} = \frac{1}{G}\sum_{k=1}^G r^{k}, \epsilon=10^{-8}.
    \label{eq:standard_advantage}
\end{equation}
Here $r^{g}$ is the reward of branch $g$, $\mathrm{clip}(\cdot)$ constrains the importance ratio within a trust region, and $\pi_{\mathrm{old}}$ is updated once per optimization iteration. We adopt an asymmetric clipping range with $\epsilon_{\mathrm{low}} = 0.1$ and $\epsilon_{\mathrm{high}} = 0.2$, which more aggressively promotes the optimization of high-reward branches while conservatively suppressing low-reward ones to prevent optimization collapse.

\textbf{Derivation.} We now verify that the velocity-field surrogate policy induces the desired preference optimization direction through its gradient.

\begin{proofbox}

\textbf{Step 1:}
We start from the unclipped Policy-Gradient (PG) objective underlying PPO:
$J_{\mathrm{PG}}(\theta)
= \mathbb{E}_{g \sim \pi_{\mathrm{old}}}
\bigl[ \tfrac{\pi_\theta(g)}{\pi_{\mathrm{old}}(g)} A^{g} \bigr]$.
Taking the gradient with respect to $\theta$ gives
$\nabla_\theta J_{\mathrm{PG}}(\theta)
= \mathbb{E}_{g \sim \pi_{\mathrm{old}}}
[\tfrac{\nabla_\theta \pi_\theta(g)}{\pi_{\mathrm{old}}(g)} A^{g}]$.
Applying the log-derivative identity 
$\nabla_\theta \log \pi_\theta(g) = \frac{\nabla_\theta \pi_\theta(g)}{\pi_\theta(g)}$, we obtain
\begingroup
\setlength{\jot}{0pt}
\begin{align}
    \nabla_\theta J_{\mathrm{PG}}(\theta)
    &= \mathbb{E}_{g \sim \pi_{\mathrm{old}}}
    [\tfrac{\pi_\theta(g)}{\pi_{\mathrm{old}}(g)} A^{g} \, \nabla_\theta \log \pi_\theta(g)] \nonumber \\
    &= \sum_{g=1}^{G} \pi_{\mathrm{old}}(g)
    \tfrac{\pi_\theta(g)}{\pi_{\mathrm{old}}(g)} A^{g} \, \nabla_\theta \log \pi_\theta(g)
     = \sum_{g=1}^{G} \pi_\theta(g) \, A^{g} \, \nabla_\theta \log \pi_\theta(g).
    \label{eq:pg_expanded_trick}
\end{align}
\endgroup

\textbf{Step 2:}
From Eq.~\eqref{eq:gibbs_policy}, we have
$\log \pi_\theta(g) = -\frac{\mathcal{E}_\theta(X^{g})}{\tau} - \log \sum_h \exp\!\left(-\frac{\mathcal{E}_\theta(X^{h})}{\tau}\right)$. Therefore,
$\nabla_\theta \log \pi_\theta(g)
= -\frac{1}{\tau}
\left(
\nabla_\theta \mathcal{E}_\theta(X^{g})
- \sum_{k=1}^{G} \pi_\theta(k) \, \nabla_\theta \mathcal{E}_\theta(X^{k})
\right).$
Substituting into~\eqref{eq:pg_expanded_trick} yields
\begingroup
\setlength{\jot}{0pt}
\begin{align}
    \nabla_\theta J_{\mathrm{PG}}(\theta)
    &= -\frac{1}{\tau}
    \, \sum_{g=1}^{G} \pi_\theta(g) \, A^{g}
    (
        \nabla_\theta \mathcal{E}_\theta(X^{g})
        - \sum_{k=1}^{G} \pi_\theta(k)
        \, \nabla_\theta \mathcal{E}_\theta(X^{k})
    ) \label{eq:pg_substituted} \\
    &= -\frac{1}{\tau}
    [\sum_{g=1}^{G} \pi_\theta(g) \, A^{g} \, \nabla_\theta \mathcal{E}_\theta(X^{g})
        - (\sum_{g=1}^{G} \pi_\theta(g) \, A^{g})
          (\sum_{k=1}^{G} \pi_\theta(k) \, \nabla_\theta \mathcal{E}_\theta(X^{k}))] \nonumber \\
    &= -\frac{1}{\tau}
    \, \sum_{g=1}^{G} \pi_\theta(g)
    (A^{g} - \sum_{k=1}^{G} \pi_\theta(k) \, A^{k})
    \nabla_\theta \mathcal{E}_\theta(X^{g}).
    \label{eq:pg_covariance_form}
\end{align}
\endgroup

\textbf{Step 3: }
Using the TVE definition in Eq.~\eqref{eq:tve}, its gradient with respect to 
$\theta$ is
\vspace{-0.5em}
\begingroup
\setlength{\abovedisplayskip}{0pt}
\setlength{\belowdisplayskip}{0pt}
\setlength{\abovedisplayshortskip}{0pt}
\setlength{\belowdisplayshortskip}{0pt}
\begin{equation}
    \nabla_\theta \mathcal{E}_\theta\!\left(X^{g}\right)
    = \sum_{b \in \mathcal{B}} \sum_{s=1}^{S}
    \frac{2}{d}
    J_{v_\theta}^{\!\top}
    (v_\theta\!\left(z_{b,s}^{g},\, t_s,\, \mathcal{K}_{<b}\right)
        - \hat{u}_{b,s}^{g}),
    \label{eq:tve_grad_specific}
\end{equation}
\endgroup
where $J_{v_\theta}$ denotes the Jacobian of the velocity field 
with respect to the network parameters. Substituting 
Eq.~\eqref{eq:tve_grad_specific} into Eq.~\eqref{eq:pg_covariance_form}, we obtain
\begingroup
\setlength{\abovedisplayskip}{0pt}
\setlength{\belowdisplayskip}{0pt}
\setlength{\abovedisplayshortskip}{0pt}
\setlength{\belowdisplayshortskip}{0pt}
\begin{align}
    \nabla_\theta J_{\mathrm{PG}}(\theta)
    &= -\frac{1}{\tau}
    \sum_{b \in \mathcal{B}} \sum_{s=1}^{S}
    \sum_{g=1}^{G} \pi_\theta(g)
    (A^{g} - \mu_A^{\pi_\theta})
    \frac{2}{d}
    J_{v_\theta}^{\!\top}
    (v_\theta\!\left(z_{b,s}^{g},\, t_s,\, \mathcal{K}_{<b}\right)
        - \hat{u}_{b,s}^{g}),
    \label{eq:tve_pg_expanded}
\end{align}
\endgroup
where $\mu_A^{\pi_\theta} = \sum_{k=1}^{G} \pi_\theta(k) \, A^{k}$ denotes the policy-weighted average advantage. Equation~\eqref{eq:tve_pg_expanded} reveals that policy optimization over TVE reduces to a \textbf{reward-weighted contrastive flow-matching} objective. When branch $g$ is superior ($A^{g} > \mu_A^{\pi_\theta}$), the gradient drives parameter updates toward reducing the replay residual against $\hat{u}_{b,s}^{g}$, aligning the ODE dynamics more consistent with the high-reward trajectory. Conversely, when branch $g$ is inferior ($A^{g} < \mu_A^{\pi_\theta}$), updates are steered away from reproducing that branch's velocity targets, suppressing low-reward trajectory dynamics. This establishes TVE as a principled bridge between reinforcement learning and the generator's native flow-matching objective~\cite{lipman2023flow}.
\end{proofbox}
\vspace{-0.5em}

\subsection{Reward Design and Regularization}
\vspace{-0.5em}
\label{subsec:reward}
\textbf{Multi-reward Formulation.} To mitigate reward hacking~\cite{liu2026beyond}, we adopt a composite reward integrating three complementary dimensions: Visual Quality (VQ), Motion Quality (MQ), and Text-Video Alignment (TA). The VQ reward is computed as the average HPSv3 score~\cite{hpsv3}, while MQ and TA rewards are obtained via the official VideoAlign configuration~\cite{videoalign}. For long-video generation, rewards are computed per segment and averaged across segments.

\textbf{KL Regularization.} To prevent the surrogate policy from drifting excessively from the pretrained distribution, we augment the objective with a discrete KL divergence penalty:
\begin{equation}
    \mathcal{D}_{\text{KL}}(\pi_\theta \| \pi_{\mathrm{ref}}) = \sum_{g=1}^G \pi_\theta(g) \left[ \log \pi_\theta(g) - \log \pi_{\mathrm{ref}}(g) \right].
    \label{eq:kl}
\end{equation}
Here $\pi_{\mathrm{ref}}$ denotes the frozen reference policy constructed with the same surrogate mapping. The total training objective combines the PPO loss (Eq.~\ref{eq:ppo_objective}) with the KL penalty (Eq.~\ref{eq:kl}):
\begin{equation}
    \mathcal{L}_{\text{total}} = \mathcal{L}_{\text{PPO}} + \beta \, \mathcal{D}_{\text{KL}}(\pi_\theta \| \pi_{\mathrm{ref}}),
\end{equation}
where $\beta$ controls the KL penalty strength. To guard against occasional pathological exploration causing model degradation, \KVPO{} zeros out the gradient for any iteration in which no candidate branch reward exceeds the anchor reward $r^{0}$.

\vspace{-0.5em}
\begin{figure*}[t]
    \centering
     \setlength{\abovecaptionskip}{-1.25em}   
    \includegraphics[width=\textwidth]{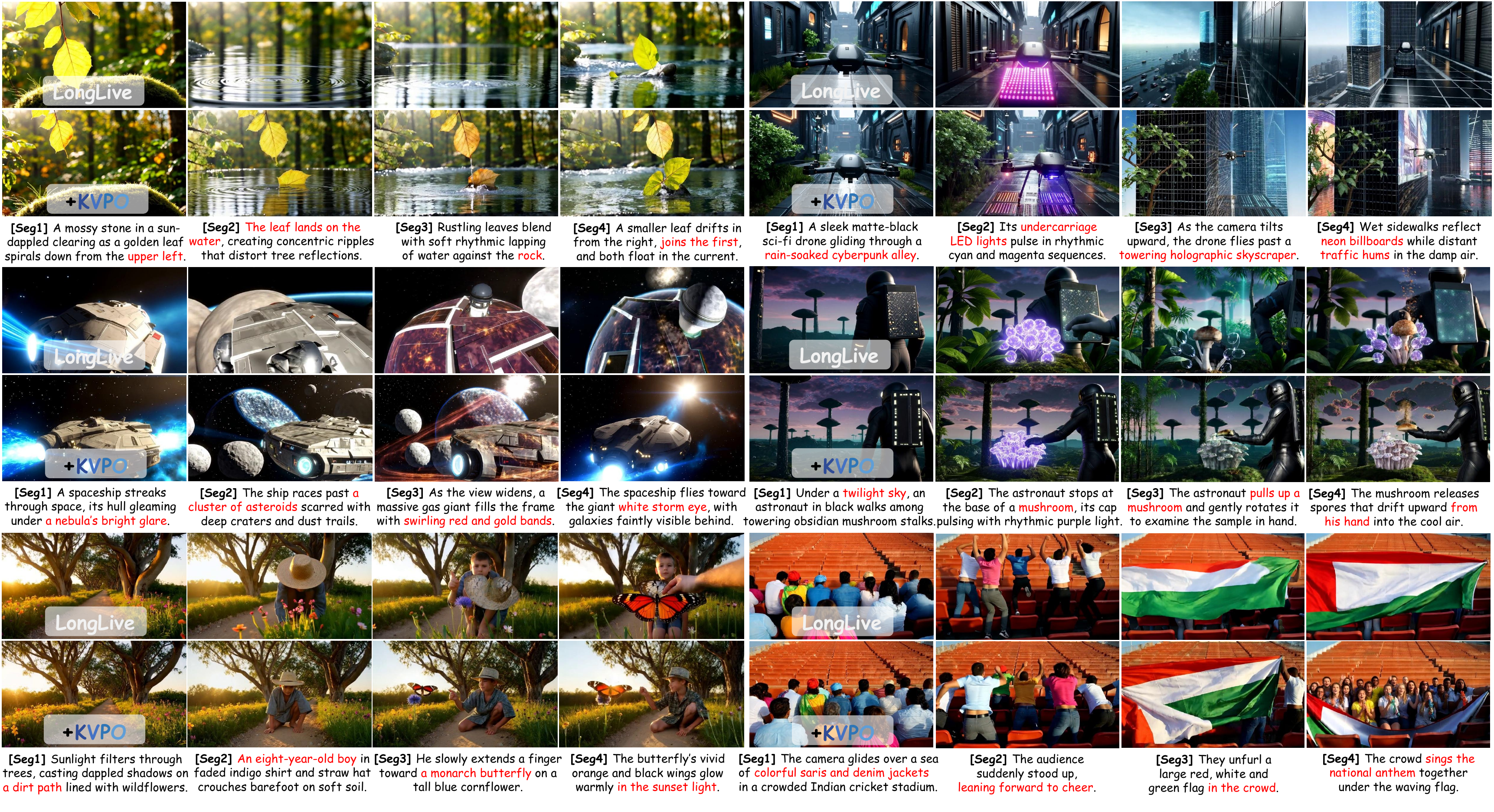}
    \caption{\textbf{Qualitative comparison between LongLive and LongLive trained with \KVPO{}}. \KVPO{} yields more faithful prompt grounding, cleaner object interactions, and smoother temporal evolution.}
    \label{fig:longlive_vis}
    \vspace{-1em}
\end{figure*}

\section{Experiments}
\vspace{-0.5em}
\subsection{Experimental Setup}
\vspace{-0.5em}
\textbf{Implementation Details.} We evaluate \KVPO{} on two state-of-the-art autoregressive video generators, LongLive~\cite{longlive} and MemFlow~\cite{memflow}. Both are obtained via classical Self-Forcing-style~\cite{selfforcing} distillation and support single-prompt and multi-prompt generation. We also compare against Astrolabe~\cite{astrolabe}, a state-of-the-art post-training method for AR video generation. Training prompts are sampled from the multi-prompt VidProM dataset~\cite{vidprom} and further refined using Qwen3~\cite{qwen3}. Each video is uniformly segmented into groups of four prompts, with prompt switching every 588 frames (147 latent frames). For parameter-efficient fine-tuning, we apply LoRA~\cite{lora} with rank \( r = 256 \) and scaling factor \( \alpha = 256 \). All experiments are conducted on 32 NVIDIA H200 GPUs, where each training iteration processes 32 prompts with a candidate group size of \( G = 8 \). Each iteration takes approximately 960 seconds, and the best checkpoint typically emerges within 3,000--4,000 training samples, corresponding to roughly 30 hours of wall-clock time and about 1000 GPU-hours. Additional key training hyperparameters are summarized in Appendix~\ref{app:kvpo_hparams}.

\begin{figure*}[t]
    \centering
     \setlength{\abovecaptionskip}{-1.25em}   
    \includegraphics[width=\textwidth]{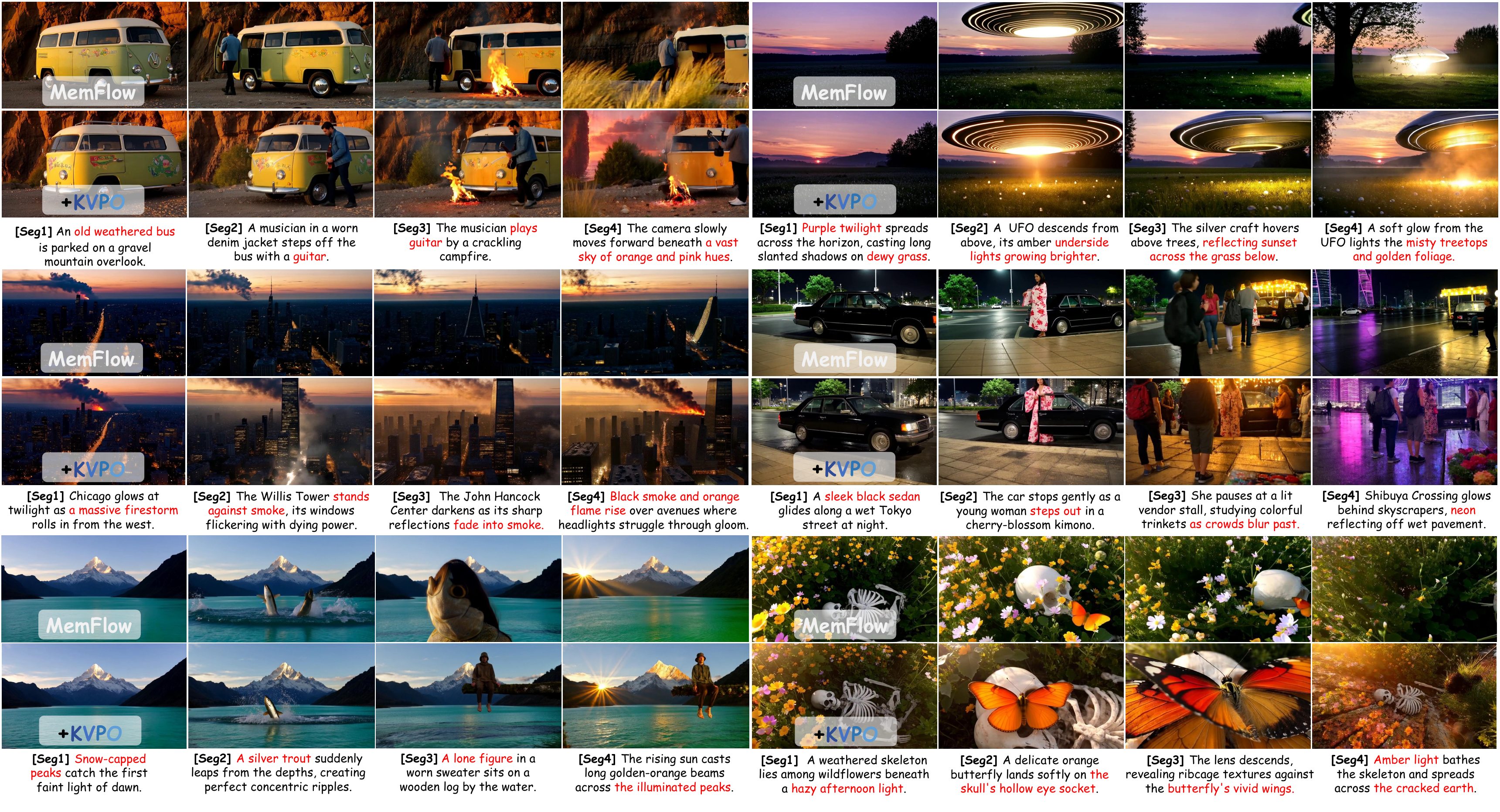}
    \caption{\textbf{Qualitative comparison between MemFlow and MemFlow trained with \KVPO{}.} \KVPO{} yields clearer semantic transitions, richer details, and stronger cross-segment consistency. }
    \label{fig:memflow_vis}
    \vspace{-0.5em}
\end{figure*}

\textbf{Evaluation Metrics.} We evaluate \KVPO{} under two settings: single-prompt short-video and multi-prompt long-video generation. In addition to the three primary metrics used by our reward design, we report four complementary VBench~\cite{vbench2} metrics, namely Quality, Semantic, Consistency Score, and CLIP Score, to provide a comprehensive assessment of model performance.

\definecolor{tablegray}{RGB}{242,242,242}
\definecolor{tableblue}{RGB}{226,238,252}

\begin{table}[t]
\centering
\tiny
\caption{\textbf{Comparison of single-prompt short-video and multi-prompt long-video generation.}}
\vspace{-0.5em}
\label{tab:combined_results}
\setlength{\tabcolsep}{3.3pt}
\renewcommand{\arraystretch}{0.8}
\resizebox{0.98\columnwidth}{!}{%
\begin{tabular}{lccccccc}
\toprule
\textbf{Method} & \textbf{VQ$\uparrow$} & \textbf{MQ$\uparrow$} & \textbf{TA$\uparrow$} & \textbf{Quality$\uparrow$} & \textbf{Semantic$\uparrow$} & \textbf{Consistency Score$\uparrow$} & \textbf{CLIP Score$\uparrow$} \\
\midrule
\rowcolor{tablegray!85}\multicolumn{8}{c}{\fontsize{6.6}{7.2}\selectfont\textbf{Single-prompt short-video generation}} \\
\midrule
\textbf{LongLive}~\cite{longlive} & 8.86 & 1.80 & 0.02 & \textbf{81.89} & 70.10 & 89.12 & 32.01 \\
\quad + Astrolabe & 9.98 & 1.87 & 0.03 & 81.41 & 70.61 & 89.14 & \textbf{32.31} \\
\rowcolor{tableblue}
\quad + \KVPO{} & \textbf{10.21}~\upimprove{15.2} & \textbf{1.89}~\upimprove{5.0} & \textbf{0.06}~\upimprove{200.0} & 81.44 & \textbf{71.45} & \textbf{89.56} & 32.29 \\
\textbf{MemFlow}~\cite{memflow} & 8.83 & 1.82 & 0.02 & 80.72 & 71.31 & 88.74 & 31.96 \\
\quad + Astrolabe & 9.47 & 1.85 & 0.03 & 80.13 & 71.42 & 88.87 & 32.04 \\
\rowcolor{tableblue}
\quad + \KVPO{} & \textbf{9.71}~\upimprove{9.1} & \textbf{1.87}~\upimprove{2.7} & \textbf{0.03}~\upimprove{50.0} & \textbf{80.91} & \textbf{71.65} & \textbf{89.08} & \textbf{32.17} \\
\midrule
\rowcolor{tablegray!85}\multicolumn{8}{c}{\fontsize{6.6}{7.2}\selectfont\textbf{Multi-prompt long-video generation}} \\
\midrule
LongLive~\cite{longlive} & 6.34 & 1.41 & -0.19 & 78.42 & 67.88 & 88.37 & 31.90 \\
\quad + Astrolabe & 7.26 & 1.44 & -0.18 & 78.46 & 68.36 & 88.30 & 32.18 \\
\rowcolor{tableblue}
\quad + \KVPO{} & \textbf{8.14}~\upimprove{28.4} & \textbf{1.50}~\upimprove{6.4} & \textbf{-0.14}~\upimprove{26.3} & \textbf{79.31} & \textbf{69.02} & \textbf{88.62} & \textbf{32.29} \\
MemFlow~\cite{memflow} & 6.30 & 1.39 & -0.20 & 77.95 & 68.11 & 87.34 & 31.80 \\
\quad + Astrolabe & 6.52 & 1.35 & -0.23 & 78.02 & 67.94 & 87.35 & 31.86 \\
\rowcolor{tableblue}
\quad + \KVPO{} & \textbf{6.96}~\upimprove{10.5} & \textbf{1.44}~\upimprove{3.6} & \textbf{-0.17}~\upimprove{15.0} & \textbf{78.36} & \textbf{68.74} & \textbf{87.52} & \textbf{32.34} \\
\bottomrule
\end{tabular}%
}
    \vspace{-2em}
\end{table}

\subsection{Quantitative and Qualitative Results}
\vspace{-0.5em}

\begin{wrapfigure}{r}{0.63\textwidth}
    \vspace{-1.5em}
    \centering
    \setlength{\abovecaptionskip}{-1.0em}
    \includegraphics[width=0.63\textwidth]{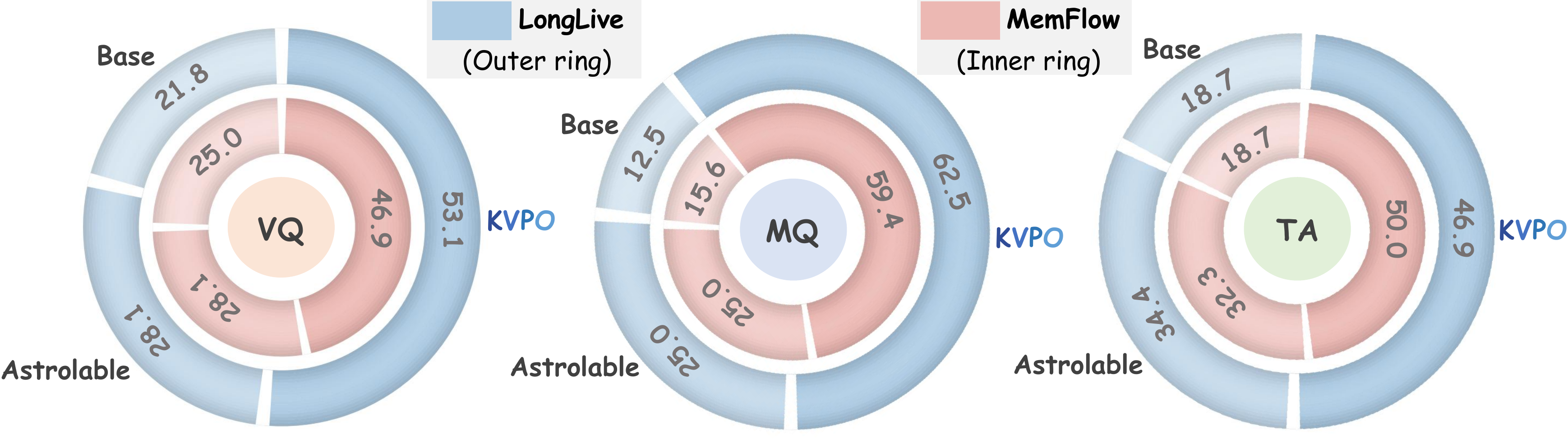}
    \caption{\textbf{Human study on long-video settings.}}
    \label{fig:user_study}
    \vspace{-1.5em}
\end{wrapfigure}

As shown in Table~\ref{tab:combined_results}, \KVPO{} achieves consistent improvements across all reward and VBench metrics in both single-prompt short-video and multi-prompt long-video settings. In the short-horizon setting, \KVPO{} improves LongLive by 15.2\%, 5.0\%, and 200.0\% on VQ, MQ, and TA respectively, and MemFlow by 9.1\%, 2.7\%, and 50.0\%. In the long-horizon setting, LongLive improves by 28.4\%, 6.4\%, and 26.3\%, while MemFlow improves by 10.5\%, 3.6\%, and 15.0\% on the same metrics. \KVPO{} also consistently outperforms Astrolabe~\cite{astrolabe}, with the margin widening in the multi-prompt long-video setting. We attribute this to causal-semantic exploration, which yields richer and semantically coherent optimization signals that better guide storyline evolution, whereas Astrolabe's noise-based exploration primarily affects low-level appearance rather than high-level semantic development.

\begin{wrapfigure}{r}{0.63\textwidth}
    \vspace{-2.25em}
    \centering
    \setlength{\abovecaptionskip}{-1.0em}
    \includegraphics[width=0.63\textwidth]{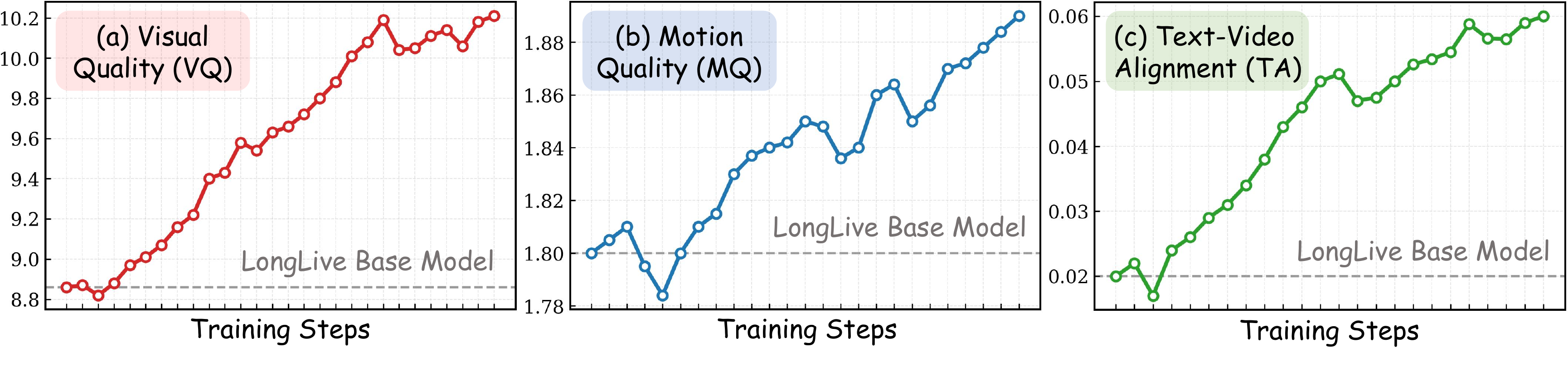}
    \caption{\textbf{Performance improvements on LongLive.}}
    \label{fig:train_plt}
    \vspace{-1.5em}
\end{wrapfigure}

Figures~\ref{fig:longlive_vis} and~\ref{fig:memflow_vis} visually confirm these improvements. Relative to the baselines, the optimized models exhibit more accurate prompt grounding, cleaner object boundaries, more plausible motion continuity, and better preservation of subject identity across segments, with fewer abrupt semantic shifts and less structural drift under viewpoint and prompt changes.
We further evaluate perceptual quality via a human study in which 32 instructed participants performed preference voting among the baseline, Astrolabe~\cite{astrolabe}, and \KVPO{} across VQ, MQ, and TA. Figure~\ref{fig:user_study} reports the proportion of samples voted best by participants under each metric, where \KVPO{} secures a clear majority preference over all competing methods.
We attribute these advantages to two design choices. Causal-Semantic Exploration enables semantically meaningful variation over causal history, improving storyline coherence in a way prior post-training methods struggle to achieve. The Velocity-Field Surrogate Policy provides an ODE-native optimization signal aligned with the generator's intrinsic flow-matching dynamics, more effectively embedding human preferences into the model. Furthermore, Appendix~\ref{app:add_qualitative} provides additional qualitative results.

\vspace{-0.5em}
\subsection{Ablation Studies}
\vspace{-0.5em}

 \begin{wrapfigure}{r}{0.56\textwidth}
    \vspace{-2.5em}
    \centering
    \small
    \captionof{table}{\textbf{Ablation of CHR and surrogate policy on LongLive in the multi-prompt long-video setting.}}
    \label{tab:ablation}
    \setlength{\tabcolsep}{3.0pt}
    \renewcommand{\arraystretch}{0.93}
    \resizebox{0.56\textwidth}{!}{%
    \begin{tabular}{lcccc}
        \toprule
        Factor & Variant & VQ$\uparrow$ & MQ$\uparrow$ & TA$\uparrow$ \\
        \midrule
        Perturbed blocks & 3 & 6.92 & 1.43 & -0.18 \\
        Perturbed blocks & \textbf{5} & \textbf{8.14} & 1.50 & \textbf{-0.14} \\
        Perturbed blocks & 7 & 8.10 & \textbf{1.53} & -0.16 \\
        \midrule
        Perturbed local KV slots & 3 / 9 & 6.22 & 1.36 & -0.20 \\
        Perturbed local KV slots & \textbf{6 / 9} & \textbf{8.14} & \textbf{1.50} & \textbf{-0.14} \\
        Perturbed local KV slots & 9 / 9 & 6.97 & 1.43 & -0.17 \\
        \midrule
        Local KV length & \textbf{Fixed 9} & \textbf{8.14} & \textbf{1.50} & \textbf{-0.14} \\
        Local KV length & Random $\{6,9,12\}$ & 8.11 & 1.48 & -0.15 \\
        \midrule
        Perturbed solver steps & 1 & 7.12 & 1.43 & -0.18 \\
        Perturbed solver steps & \textbf{2} & \textbf{8.14} & 1.50 & -0.14 \\
        Perturbed solver steps & 3 & 7.65 & \textbf{1.51} & \textbf{-0.12} \\
        Perturbed solver steps & 4 & 7.41 & 1.46 & -0.17 \\
        \midrule
        Surrogate policy & Geometric latent $\ell_2$ & 6.02 & 1.43 & -0.21 \\
        Surrogate policy & \textbf{TVE} & \textbf{8.14} & \textbf{1.50} & \textbf{-0.14} \\
        \bottomrule
    \end{tabular}%
    }
    \vspace{-1.5em}
\end{wrapfigure}

Table~\ref{tab:ablation} presents a comprehensive ablation of the core components of \KVPO{} on LongLive~\cite{longlive} under the multi-prompt long-video setting, with additional ablations provided in Appendix~\ref{app:ablation}.

\textbf{Causal History Routing (CHR).} We ablate the main CHR hyperparameters, including the number of perturbed blocks, perturbed local KV slots, local KV window length, and perturbed denoising steps. First, perturbing $5$ blocks yields the best overall trade-off, since using only $3$ blocks likely provides insufficient semantic variation and weakens all metrics, while $7$ blocks offer no consistent gain while incurring higher memory cost. Second, perturbing $6$ of the $9$ local KV slots is optimal. Too few perturbed slots produce overly similar branches and thus weak preference signals, whereas perturbing all $9$ disrupts short-range temporal anchoring and destabilizes generation. Third, randomizing the local KV window length brings negligible improvement, suggesting that the default local window already balances causal context and exploration diversity while remaining matched to inference. Fourth, perturbing the first two denoising steps yields the most balanced improvement across all three metrics. Perturbing only one step provides insufficient intervention on coarse semantics and motion layout, whereas perturbing additional steps significantly degrades visual quality and substantially increases memory overhead.

\textbf{TVE-based Surrogate Policy.} Replacing TVE with a geometric latent-space $\ell_2$ surrogate, similar to that used in NeighborGRPO~\cite{he2025neighbor}, substantially degrades performance across all metrics, demonstrating that velocity-field-space policy modeling is critical for effective preference optimization under autoregressive ODE replay. We further analyze the limitations of Euclidean-distance surrogate policies and the advantages of our formulation in the Appendix~\ref{subsec:theoretical_justification}.

\vspace{-1em}
\section{Conclusion}
\vspace{-1em}
We studied preference alignment for streaming autoregressive video generators under the deterministic ODE regime. To address the mismatch between existing noise-driven reinforcement learning methods and ODE-based generation, we introduced \KVPO{}, which combines causal-semantic exploration through Causal History Routing (CHR) with a velocity-field surrogate policy based on Trajectory Velocity Energy (TVE). CHR redirects exploration to the historical key-value cache, inducing semantically meaningful and on-manifold candidate branches, while the TVE-based surrogate policy keeps preference optimization within the model's native flow-matching dynamics.
Experiments on distilled autoregressive video generators demonstrate consistent improvements in visual quality, motion quality, and text-video alignment across both short-video and long-video generation settings.
More broadly, \KVPO{} suggests that semantic-space exploration offers a principled alternative to noise-based perturbation for inducing on-manifold branch diversity, and that velocity-field space provides a natural domain for surrogate policy modeling faithful to the generator's intrinsic dynamics. These insights may inform future alignment research on ODE-based generative architectures.





\bibliographystyle{splncs04}
\bibliography{main}


\clearpage
\appendix
\AtBeginEnvironment{equation}{\large}
\AtBeginEnvironment{align}{\large}
\AtBeginEnvironment{gather}{\large}
\AtBeginEnvironment{multline}{\large}

\begin{center}
    {\LARGE \textbf{Appendix}}
\end{center}

\section{\KVPO{} Training Pipeline}
\label{subsec:kvpo_algorithm}

We summarize the full training procedure of \KVPO{}. At a high level, each iteration consists of three tightly coupled
stages: semantic exploration, replay-based surrogate policy construction, and policy optimization. 
\KVPO{} first shares the initial latent $x_T$ and perturbs the composition of the local KV cache to generate a group of semantically diverse candidate branches through Causal History Routing (CHR). These branches are then evaluated by the reward model to produce group advantages. The same trajectories are replayed under the unperturbed context to compute Trajectory Velocity Energy (TVE), from which \KVPO{} constructs the Gibbs-form surrogate policy and the PPO importance ratios. Finally, the generator is updated using the clipped PPO objective together with KL regularization toward the reference policy. Algorithm~\ref{alg:kvpo} gives the complete procedure.

\begin{algorithm}[ht]
    \LinesNotNumbered
    \small
    \SetAlgoLined
    \SetAlgoSkip{}
    \SetInd{0.8em}{1.4em}
    \caption{\KVPO{} Training}
    \label{alg:kvpo}
    \KwIn{Generator $v_\theta$, old policy $\pi_{\mathrm{old}}$, reference policy $\pi_{\mathrm{ref}}$, reward model $R$, prompt $\mathcal{C}$, latent $x_T$, history cache $\mathcal{K}_{<b}$, branch number $G$, solver steps $S$, perturbation window $\mathcal{B}$, PPO clips $\epsilon_{\mathrm{low}},\epsilon_{\mathrm{high}}$, KL weight $\beta$, temperature $\tau$}
    \KwOut{Optimized generator $v_{\theta^*}$}
    \For{iteration $n = 1, 2, \ldots$}{
        \tcp{\scriptsize Step 1: CHR-based semantic exploration}
        Sample pivot block $b^*$ and anchor video $X^{0} \gets \Phi(x_T;\,\theta,\mathcal{K}_{<b})$\;
        \For{$g = 1, \dots, G$}{
            Sample CHR refill sets $\{\mathcal{J}_{m}^{g}\}_{m=1}^{6}$ and form
            $\tilde{\mathcal{K}}_{<b^*}^{g,\mathrm{local}} \gets \mathrm{CHR}(\mathcal{K}_{<b^*}^{\mathrm{hist}};\,\mathcal{J}_{1:6}^{g})$\;
            Generate branch $X^{g} \gets \Phi(x_T;\,\theta,\tilde{\mathcal{K}}_{<b^*}^{g,\mathrm{local}},\mathcal{B})$\;
        }

        \tcp{\scriptsize Step 2: Cache rollout states for replay}
        For each branch $g$ and replay step $(b,s) \in \mathcal{B} \times [S]$, cache
        $\mathcal{T}^{g} \gets \mathcal{T}^{g} \cup \{(z_{b,s}^{g},\hat{u}_{b,s}^{g})\}$, then restore the default local cache after $\mathcal{B}$\;

        \tcp{\scriptsize Step 3: Compute group-normalized advantages}
        Evaluate rewards $r^{g} \gets R(X^{g},\mathcal{C})$ for all $g$\;
        Compute $\bar{r} \gets \frac{1}{G}\sum_{k=1}^{G} r^{k}$ and
        $\sigma_r \gets \sqrt{\frac{1}{G}\sum_{k=1}^{G}(r^{k}-\bar{r})^2}+\epsilon$\;
        Set $A^{g} \gets \frac{r^{g}-\bar{r}}{\sigma_r}$ for each branch $g$\;

        \tcp{\scriptsize Step 4: Replay and construct the Gibbs surrogate}
        Replay all branches under the shared unperturbed context $\mathcal{K}_{<b}$\;
        \For{$g = 1, \dots, G$}{
            $\mathcal{E}_\theta(X^{g}) \gets \sum_{b \in \mathcal{B}} \sum_{s=1}^{S}
            \frac{1}{d}\,\|v_\theta(z_{b,s}^{g},t_s,\mathcal{K}_{<b})-\hat{u}_{b,s}^{g}\|_F^2$\;
            $\ell_\theta^{g} \gets -\mathcal{E}_\theta(X^{g})/\tau$,\quad
            $\pi_\theta(g) \gets \frac{\exp(\ell_\theta^{g})}{\sum_{h=1}^{G}\exp(\ell_\theta^{h})}$,\quad
            $\rho^{g} \gets \frac{\pi_\theta(g)}{\pi_{\mathrm{old}}(g)}$\;
        }

        \tcp{\scriptsize Step 5: PPO-KL update}
        $\mathcal{L}_{\mathrm{PPO}} \gets -\frac{1}{G}\sum_{g=1}^{G}
        \min\!\Big(\rho^{g}A^{g},\,\mathrm{clip}(\rho^{g},1-\epsilon_{\mathrm{low}},1+\epsilon_{\mathrm{high}})A^{g}\Big)$\;
        $\mathcal{D}_{\mathrm{KL}} \gets \sum_{g=1}^{G}\pi_\theta(g)
        \big[\log\pi_\theta(g)-\log\pi_{\mathrm{ref}}(g)\big]$\;
        $\mathcal{L}_{\mathrm{total}} \gets \mathcal{L}_{\mathrm{PPO}} + \beta\,\mathcal{D}_{\mathrm{KL}}$\;
        $\theta \gets \theta - \eta\,\nabla_\theta \mathcal{L}_{\mathrm{total}}$,\quad
        $\pi_{\mathrm{old}} \gets \pi_\theta$\;
    }
    \Return{$v_{\theta^*}$}
\end{algorithm}

\section{Why KV Exploration?}
\label{app:kv_exploration}
We clarify why the historical KV cache is a particularly suitable locus for exploration. For streaming autoregressive video generation, an effective exploration locus should satisfy three criteria: it should meaningfully influence future generation, remain compatible with the model's native causal generation pathway, and induce diversity at the level most relevant to preference optimization. The historical KV cache is well aligned with these criteria. First, because future content is strongly conditioned on accumulated historical context, modifying routed local KV memory directly affects subsequent continuation. Second, this intervention operates on the model's native conditioning state rather than introducing external stochastic perturbations into the latent trajectory. Third, because preferences for long-horizon video depend heavily on temporal coherence, subject consistency, and semantic progression, varying historical KV composition is better suited to induce semantically distinct branches than perturbations that primarily alter local appearance.

\begin{figure*}[t]
    \centering
    \setlength{\abovecaptionskip}{-1.0em}
    \includegraphics[width=\textwidth]{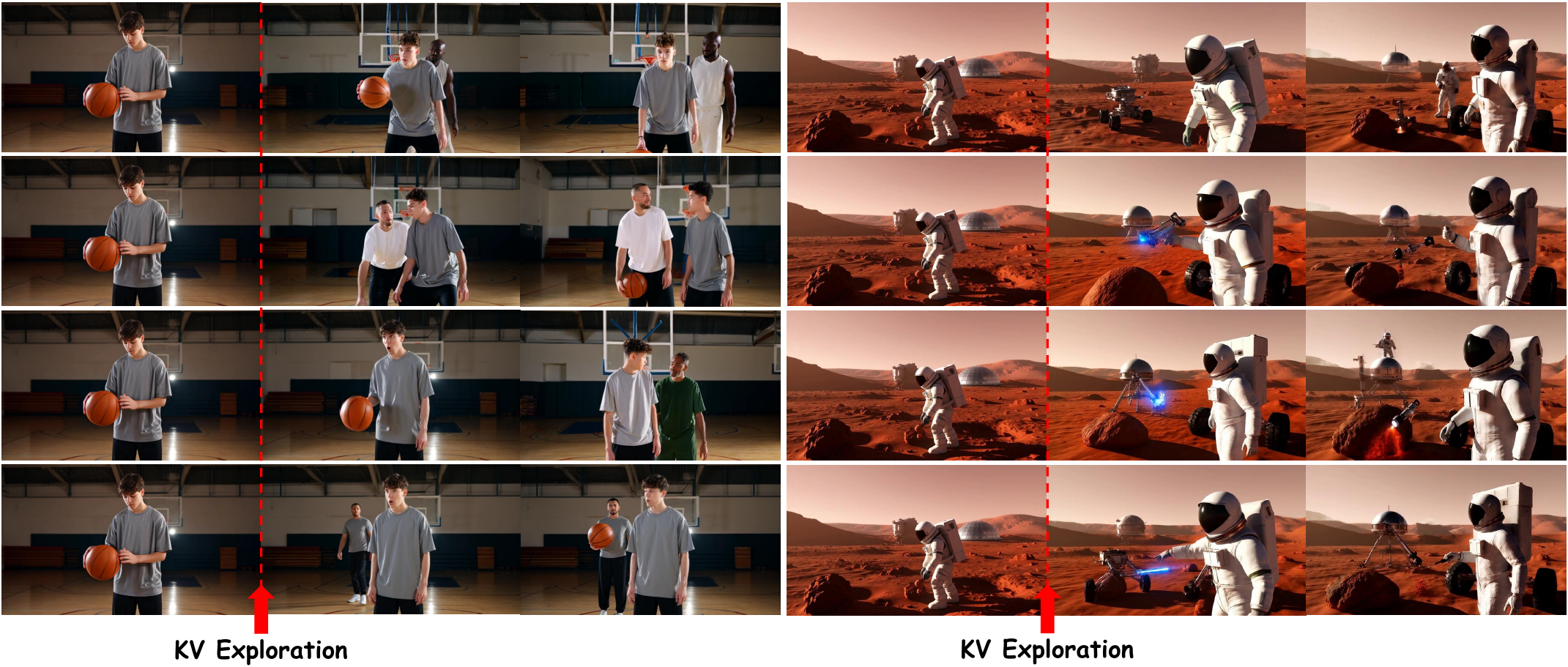}
    \caption{\textbf{Visualization of the effectiveness of causal-semantic exploration.} }
    \label{fig:kv-explore}
\end{figure*}

\section{Conditional Marginal Preservation}
We show that CHR preserves the model's conditional marginal distribution. For an ODE-based generator, this means that, for any fixed conditioning context $\mathcal{K}$, the deterministic generation map transports the base noise distribution to the corresponding conditional model distribution.

Under the probability-flow ODE framework~\cite{song2021score}, if generation is written as
\begin{equation}
    x_0 = \Phi(x_T;\,\theta,\mathcal{K}), \qquad x_T \sim \mathcal{N}(0,\mathbf{I}),
\end{equation}
then, for fixed $\mathcal{K}$, the induced sample distribution of $x_0$ is exactly the model's conditional distribution $p_\theta(x_0 \mid \mathcal{K})$. In this sense, the conditional marginal is preserved because sampling is still performed by solving the original deterministic ODE from a Gaussian initial latent.

The role of CHR is therefore not to modify this noise-to-sample transport itself, but only to change the conditioning context. Since CHR leaves the initial latent $x_T$ unchanged, each branch is generated as
\begin{equation}
    x_0^{g} = \Phi(x_T;\,\theta,\tilde{\mathcal{K}}_{<b^*}^{g,\mathrm{local}}),
\end{equation}
and thus remains an exact sample from the corresponding conditional model distribution
\begin{equation}
    x_0^{g} \sim p_\theta\bigl(x_0 \mid \tilde{\mathcal{K}}_{<b^*}^{g,\mathrm{local}}\bigr).
\end{equation}
Accordingly, CHR changes the branchwise conditioning context and hence the semantic trajectory, but it does not introduce an off-manifold perturbation to the underlying generative process.

\section{Rethinking ODE-based Policy Optimization}
\label{subsec:theoretical_justification}
We discuss and compare \KVPO{} with recent purely ODE-based policy optimization methods. Recent advances in purely ODE-based GRPO optimization (e.g., NeighborGRPO~\cite{he2025neighbor}) construct an exploration neighborhood by perturbing the initial latent and approximate the surrogate policy via Euclidean distance between generated latents. In this section, we analyze the theoretical limitations of such noise-driven exploration in autoregressive video generation and highlight the advantages of \KVPO{} from two perspectives: diversity exploration and surrogate policy modeling.

\subsection{Diversity Exploration: Disentangling Policy from Noise Variance}
Diverse candidate samples are a prerequisite for effective policy optimization. Let the generation process be denoted by $x_0 = \Phi(x_T; \theta, \mathcal{K})$. In noise-driven exploration, candidate samples are generated from a perturbed initial latent $x_T^{i} = x_T^* + \sigma \delta^{i}$, where $\delta^{i} \sim \mathcal{N}(0, \mathbf{I})$.
Applying a first-order Taylor expansion with respect to the initial latent, the variation in the generated sample $\Delta x_0^{i} = x_0^{i} - x_0^*$ can be approximated as:
\begin{equation}
    \Delta x_0^{i} \approx \nabla_{x_T} \Phi(x_T^*; \theta, \mathcal{K}) \cdot (\sigma \delta^{i})
\end{equation}
When prior works model the surrogate policy $\pi_\theta(i)$ using the Euclidean distance $\|\Delta x_0^{i}\|_2^2$, a fundamental confounding issue arises. The variation in $\Delta x_0^{i}$ is inherently coupled with the random noise seed $\delta^{i}$. Therefore, a larger distance does not necessarily imply a lower likelihood under the current surrogate policy $\theta$; it may simply correspond to a larger sampled noise magnitude. This conflation corrupts the true policy gradient signal.

\textbf{Advantage of Causal-Semantic Exploration.} \KVPO{} resolves this ambiguity by shifting exploration from noise perturbation to causal-semantic exploration over the historical KV cache. Concretely, it keeps the initial latent $x_T$ fixed across all candidates, while CHR instantiates the branch-specific local memory $\tilde{\mathcal{K}}_{<b^*}^{g,\mathrm{local}}$. Because the initial latent is shared, any structural or semantic difference among the generated trajectories can be attributed solely to the model's deterministic response to the distinct context. This disentanglement ensures that the exploration space directly reflects the surrogate preference ordering induced by changes in the historical KV cache rather than by differences in noise magnitude.

\subsection{Surrogate Policy Modeling: Overcoming Geometric Distortion}
\label{sec:surrogate policy comparison}
Modeling the surrogate policy via unweighted Euclidean distance intrinsically imposes an isotropic assumption, treating every feature dimension as contributing equally to likelihood evaluation:
\begin{equation}
    \pi_{\mathrm{noise}}(i) \propto \exp\left(-\frac{1}{2\sigma_\pi^2}\|x_0^{i} - x_0^*\|_2^2\right)
\end{equation}
This implies a local covariance of $\Sigma_{\mathrm{assume}} = \sigma_\pi^2 \mathbf{I}$.

However, the probability flow ODE $\Phi(\cdot;\theta)$ mapping $x_T$ to $x_0$ is nonlinear. By the change-of-variables theorem, the true local covariance pushed forward from the isotropic prior is governed by the Jacobian $\mathbf{J}_{\Phi}$:
\begin{equation}
    \Sigma_{\mathrm{true}} = \sigma^2 \mathbf{J}_{\Phi}(x_T^*)\mathbf{J}_{\Phi}(x_T^*)^{\top}
\end{equation}
In AR models spanning multiple continuous blocks and diffusion steps, this cascaded nonlinear transformation yields an anisotropic true distribution ($\mathbf{J}_{\Phi}\mathbf{J}_{\Phi}^{\top} \neq \mathbf{I}$). The unweighted Euclidean distance implicitly ignores this Riemannian curvature, introducing geometric distortion that unjustly penalizes valid semantic variations along the principal axes of the flow.

\textbf{Advantage of velocity-field surrogate policy modeling.} \KVPO{} resolves this by constructing the surrogate policy via trajectory replay in the velocity-field space, with TVE serving as the energy functional evaluating each candidate under the unperturbed deployment-time context:
\begin{equation}
    \pi_\theta(i) \propto \exp\left(-\mathcal{E}_\theta(X^{i} \mid \mathcal{K}_{<b}) / \tau\right)
\end{equation}
Since TVE is defined through velocity-field MSE accumulated along the replayed ODE trajectory, it directly mirrors the model's native flow-matching objective and provides a geometrically sound, algorithmically aligned gradient signal for AR video policy optimization.

\section{Ablation on the KL Penalty Weight}
\label{app:ablation}

We study the effect of the KL regularization coefficient $\beta$ in Eq.~\eqref{eq:kl}. This term anchors the learned surrogate policy to the frozen reference and is essential for stable online preference optimization. Without it ($\beta=0$), the policy drifts aggressively toward noisy high-reward branches, the PPO update becomes poorly regularized, and the replay-time distribution diverges too far from the pretrained generator — in practice, this leads to training collapse rather than meaningful reward optimization. Conversely, excessively large $\beta$ keeps training stable but renders the update too conservative. Optimal performance is thus expected at an intermediate regularization strength.

Table~\ref{tab:kl_ablation_longlive} confirms this on LongLive~\cite{longlive}. At $\beta=0$, all metrics drop substantially below the base model in both settings, indicating collapse in both alignment quality and generation stability. Introducing even a moderate KL constraint restores stable learning: $\beta=1$ recovers most of the gains, and $\beta=3$--$5$ yields the strongest overall trade-off. The default $\beta=5$ achieves the most balanced performance across reward metrics and auxiliary VBench metrics. Beyond this, gains diminish gradually as the optimization grows increasingly conservative.

\vspace{-1em}
\begin{table*}[h]
    \centering
    \tiny
    \caption{\textbf{Ablation on the KL penalty weight $\beta$ for LongLive.} }
    \label{tab:kl_ablation_longlive}
    \setlength{\tabcolsep}{3.6pt}
    \renewcommand{\arraystretch}{0.78}
    \resizebox{0.95\textwidth}{!}{%
    \begin{tabular}{lccccccc}
    \toprule
    \textbf{KL Weight $\beta$} & \textbf{VQ$\uparrow$} & \textbf{MQ$\uparrow$} & \textbf{TA$\uparrow$} & \textbf{Quality$\uparrow$} & \textbf{Semantic$\uparrow$} & \textbf{Consistency Score$\uparrow$} & \textbf{CLIP Score$\uparrow$} \\
    \midrule
    \rowcolor{tablegray!85}\multicolumn{8}{c}{\fontsize{6.3}{6.8}\selectfont\textbf{Single-prompt short-video generation}} \\
    \midrule
    \textbf{Base} & 8.86 & 1.80 & 0.02 & 81.89 & 70.10 & 89.12 & 32.01 \\
    $\beta=0$ & 6.92 & 1.36 & -0.07 & 77.42 & 65.96 & 85.84 & 30.31 \\
    $\beta=1$ & 10.04 & 1.87 & 0.07 & 81.21 & 71.19 & 89.28 & 32.23 \\
    $\beta=3$ & 10.22 & 1.88 & 0.06 & 81.37 & 71.33 & 89.46 & 32.27 \\
    \rowcolor{tableblue}
    \textbf{$\beta=5$} & 10.21 & 1.89 & 0.06 & 81.44 & 71.45 & 89.56 & 32.29 \\
    $\beta=10$ & 10.13 & 1.87 & 0.05 & 81.30 & 71.18 & 89.34 & 32.17 \\
    $\beta=20$ & 9.68 & 1.84 & 0.04 & 81.08 & 70.84 & 89.23 & 32.10 \\
    \midrule
    \rowcolor{tablegray!85}\multicolumn{8}{c}{\fontsize{6.3}{6.8}\selectfont\textbf{Multi-prompt long-video generation}} \\
    \midrule
    \textbf{Base} & 6.34 & 1.41 & -0.19 & 78.42 & 67.88 & 88.37 & 31.90 \\
    $\beta=0$ & 4.96 & 1.18 & -0.33 & 76.11 & 64.88 & 85.17 & 30.08 \\
    $\beta=1$ & 8.02 & 1.48 & -0.14 & 78.98 & 68.79 & 88.36 & 32.19 \\
    $\beta=3$ & 8.13 & 1.49 & -0.14 & 79.20 & 68.93 & 88.55 & 32.25 \\
    \rowcolor{tableblue}
    \textbf{$\beta=5$} & 8.14 & 1.50 & -0.14 & 79.31 & 69.02 & 88.62 & 32.29 \\
    $\beta=10$ & 7.96 & 1.48 & -0.15 & 79.05 & 68.84 & 88.50 & 32.18 \\
    $\beta=20$ & 7.58 & 1.45 & -0.16 & 78.73 & 68.46 & 88.34 & 32.06 \\
    \bottomrule
    \end{tabular}%
    }
    \vspace{-0.8em}
\end{table*}

\section{Limitations}
\label{app:limitations}
Although \KVPO{} shows consistent gains on the distilled autoregressive video generators studied in this paper, it still has several limitations. First, the proposed causal-semantic exploration mechanism is designed for autoregressive generators with KV-cache-based memory. While this design is natural for many mainstream autoregressive video models, extending it to models without KV caches or with substantially different memory mechanisms (e.g., Mamba-style state-space models) may require additional adaptation.
Second, although \KVPO{} introduces a novel ODE-native velocity-field surrogate for preference optimization, training incurs additional computational and memory cost due to the need to cache replay states. Although this cost grows with the number of branches, the replay horizon, and the solver steps, it remains small relative to the peak memory required for backpropagation. Finally, the optimization quality depends on the fidelity of the reward model. If the reward does not adequately capture long-range visual quality, narrative consistency, or subtle motion realism, the resulting \KVPO{} updates may not fully reflect human preferences.
Future work will focus on extending \KVPO{} to a broader range of video generation architectures and developing stronger reward models for long-horizon semantic and motion consistency.

\section{Key \KVPO{} Training Hyperparameters}
\label{app:kvpo_hparams}

\begin{figure*}[h]
    \centering
    \setlength{\abovecaptionskip}{-1.5em}
    \includegraphics[width=\textwidth]{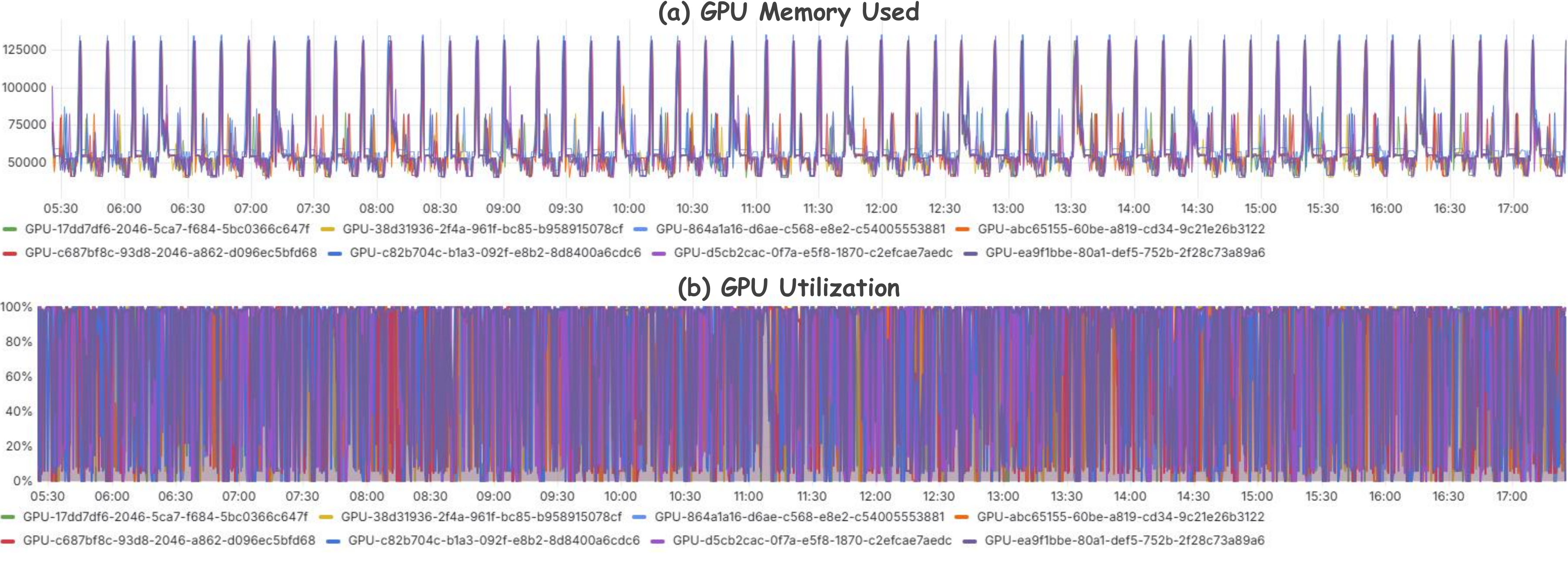}
    \caption{\textbf{GPU memory usage and compute utilization of \KVPO{} during LongLive training.}}
    \label{fig:gpu_log}
\end{figure*}

Table~\ref{tab:kvpo_hparams} summarizes the key hyperparameters used in \KVPO{} training. Figure~\ref{fig:gpu_log} reports the GPU memory footprint and compute utilization during LongLive training, with peak memory usage of approximately 130\,GB and a per-step runtime of approximately 960 seconds.

\begin{table*}[h]
    \centering
    \small
    \setlength{\tabcolsep}{4pt}
    \renewcommand{\arraystretch}{1.0}
    \caption{\textbf{Comprehensive hyperparameters for \KVPO{} training.} We summarize the configurations used for model setup, optimization, replay-based reinforcement learning, and streaming rollout.}
    \label{tab:kvpo_hparams}
    \begin{tabular}{p{2.8cm}p{4.4cm}p{5.4cm}}
        \toprule
        \textbf{Category} & \textbf{Setting} & \textbf{Value} \\
        \midrule
        \textbf{Model \& Video Specs} 
         & Denoising Timesteps $(T)$ & 4 from $[1000, 750, 500, 250]$ \\
         & Local Attention Size & 12 \\
         & Sink Size & 3 \\
         & Local Size & 9 \\
         & Frames per Block & 3 \\
        \midrule
        \textbf{LoRA Fine-Tuning} & Rank $(r)$ & 256 \\
         & Scaling Factor $(\alpha)$ & 256 \\
         & Dropout Rate & 0.0 \\
         & Gradient Checkpointing & True \\
         & EMA Decay Rate & 0.999 \\
        \midrule
        \textbf{Optimization} & Hardware & 32 NVIDIA H200 GPUs \\
         & Optimizer & AdamW \\
         & Learning Rate $(\eta)$ & $1\times10^{-5}$ \\
         & LR Scheduler & constant\_with\_warmup \\
         & Warmup Steps & 5 \\
         & Weight Decay & 0.01 \\
         & Gradient Accumulation Steps & 2 \\
         & Max Gradient Norm & 1.0 \\
         & Max Train Steps & 1500 \\
        \midrule
        \textbf{Replay-Based RL} & Branch Number $(K)$ & 8 \\
         & Perturbed Blocks $(P)$ & 5 \\
         & Nearest Preserved Frames $(m)$ & 3 \\
         & Perturb Search Range & first 7 chunks \\
         & Reward Signals & local + global \\
         & Local / Global Loss Weights & 1.0 / 0.5 \\
         & Gradient-Carrying Replay Steps & 2 \\
         & PPO Epochs & 1 \\
         & Clip Range $(\epsilon_{\mathrm{low}}, \epsilon_{\mathrm{high}})$ & (0.1, 0.2) \\
         & Advantage Clip Max & 2.5 \\
         & KL Penalty Weight & 5 \\
         & KL Reference Policy & frozen initialization policy \\
        \midrule
        \textbf{Streaming Rollout} & Chunk Size & 21 \\
         & Prompt Segments & 4 \\
         & Switch Frame Indices & [39, 75, 111] \\
         & VAE Temporal Stride & 4 \\
        \bottomrule
    \end{tabular}

\end{table*}

\clearpage

\section{More Qualitative Results}
\label{app:add_qualitative}

This section provides additional qualitative comparisons complementing the quantitative results in the main paper, covering both LongLive~\cite{longlive} and MemFlow~\cite{memflow} settings. The comparisons span diverse prompts and scene transitions, enabling evaluation of long-horizon temporal coherence, subject identity preservation, motion plausibility, and text-video alignment. Across all presented cases, \KVPO{} consistently exhibits stronger semantic consistency and more stable narrative progression than the compared baselines.

\begin{figure*}[h]
    \centering
    \includegraphics[width=\textwidth]{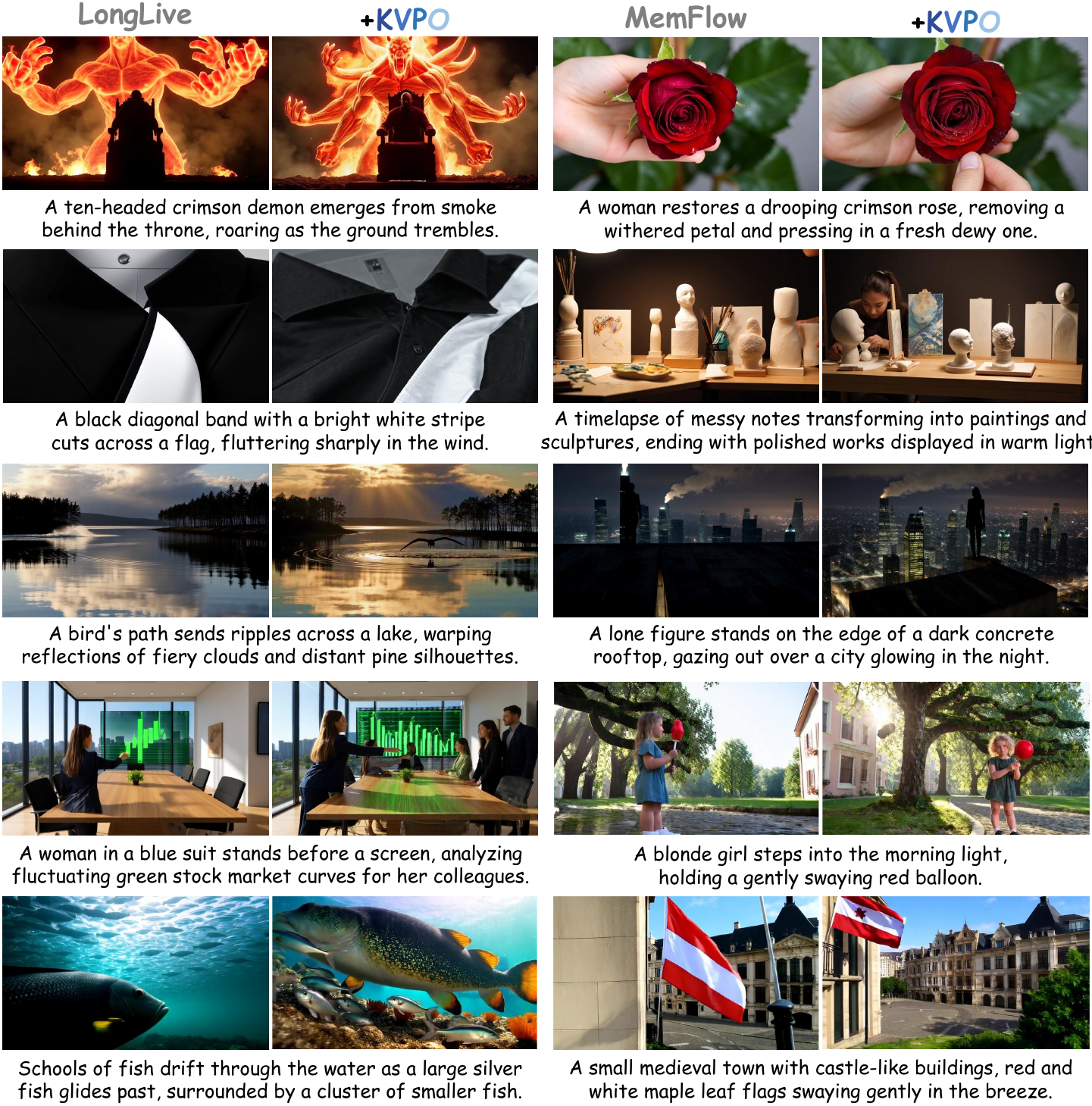}
    \caption{\textbf{Additional qualitative results on short-video generation.}}
    \label{fig:app-short-vis}
\end{figure*}

\begin{figure*}[p]
    \centering
    \includegraphics[width=\textwidth]{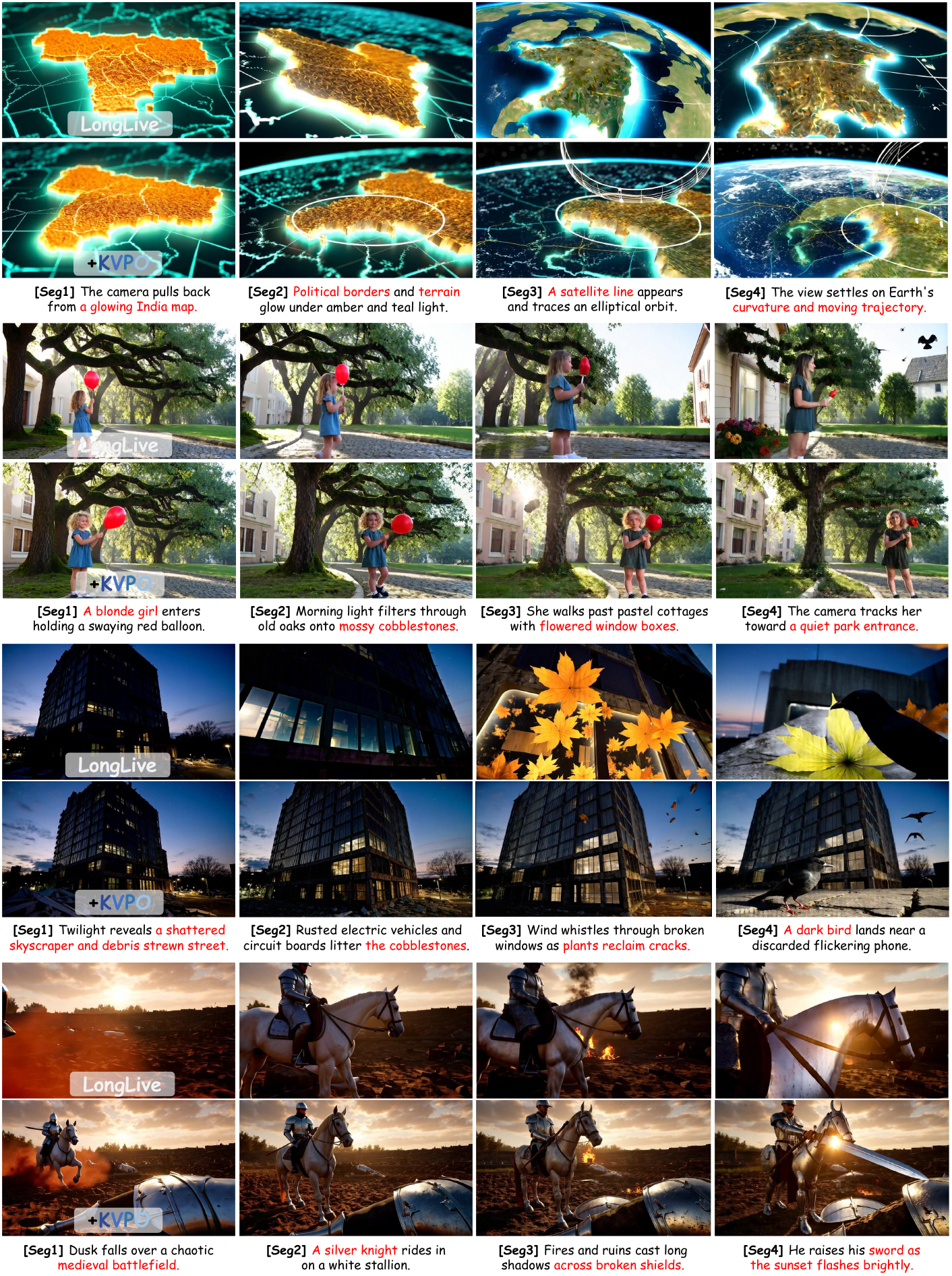}
    \caption{\textbf{Additional qualitative results on \textbf{LongLive}.}}
    \label{fig:app-longlive-1}
\end{figure*}

\begin{figure*}[p]
    \centering
    \includegraphics[width=\textwidth]{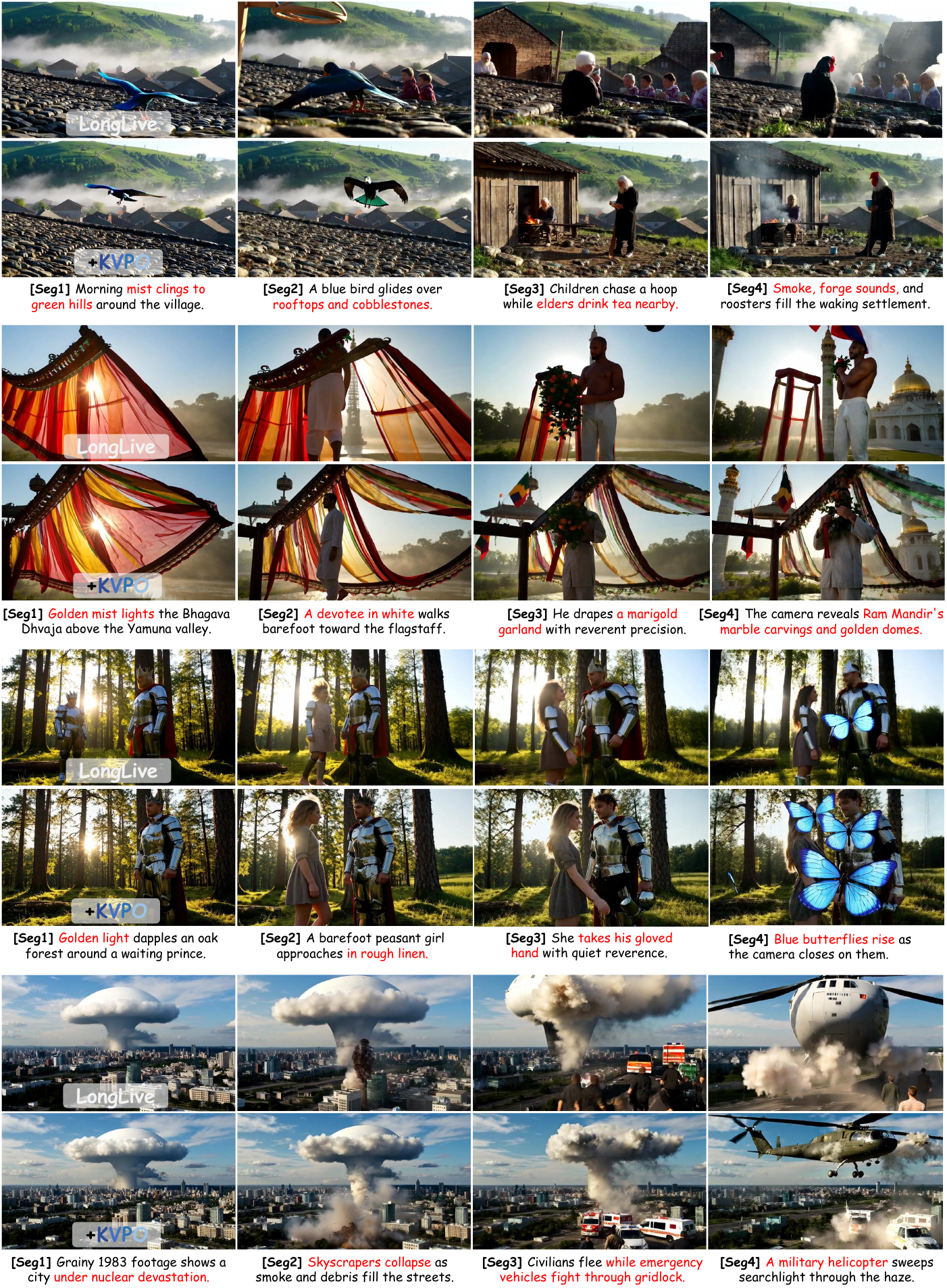}
    \caption{\textbf{Additional qualitative results on \textbf{LongLive}.}}
    \label{fig:app-longlive-2}
\end{figure*}

\begin{figure*}[p]
    \centering
    \includegraphics[width=\textwidth]{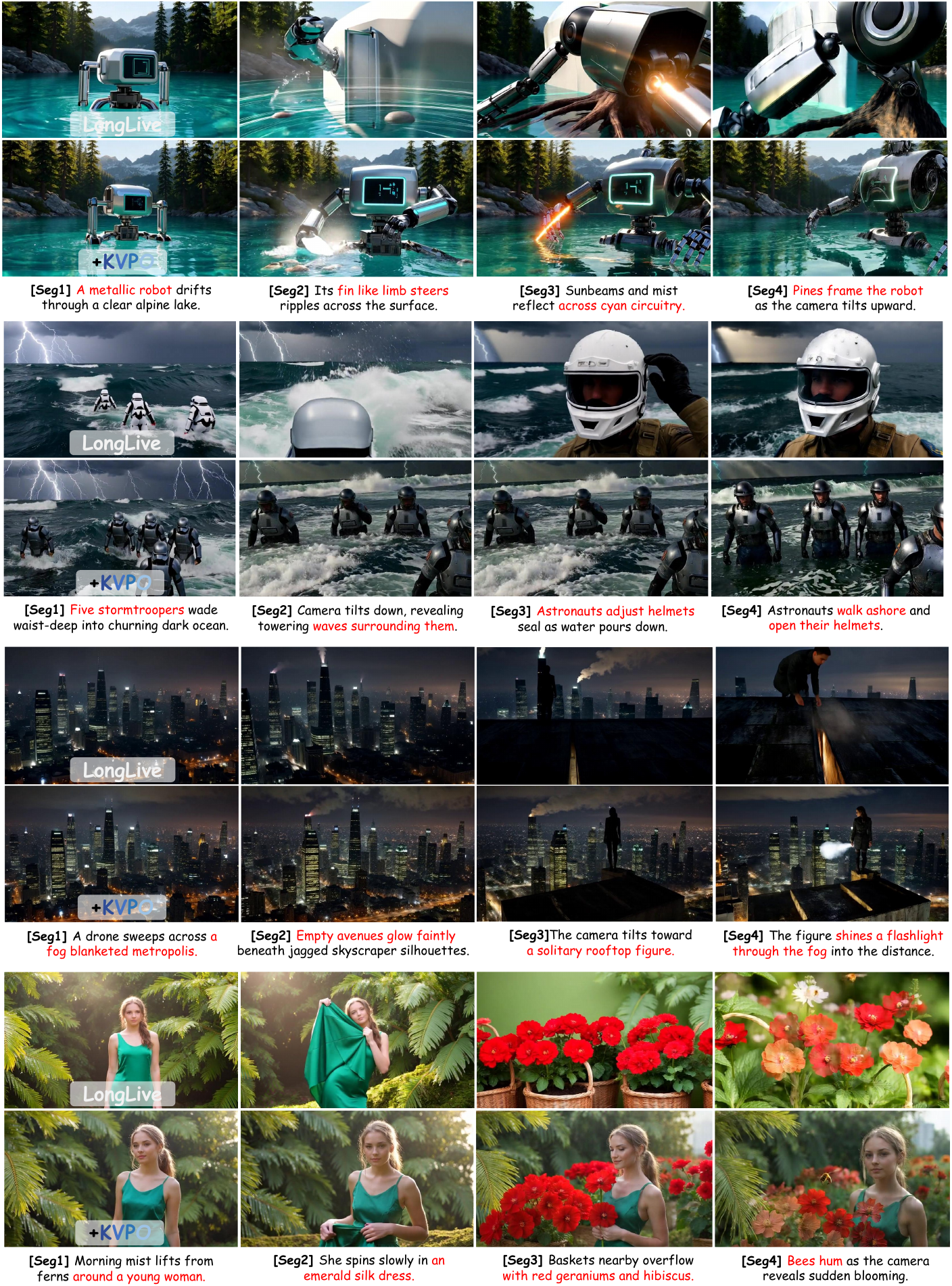}
    \caption{\textbf{Additional qualitative results on \textbf{LongLive}.}}
    \label{fig:app-longlive-3}
\end{figure*}

\begin{figure*}[p]
    \centering
    \includegraphics[width=\textwidth]{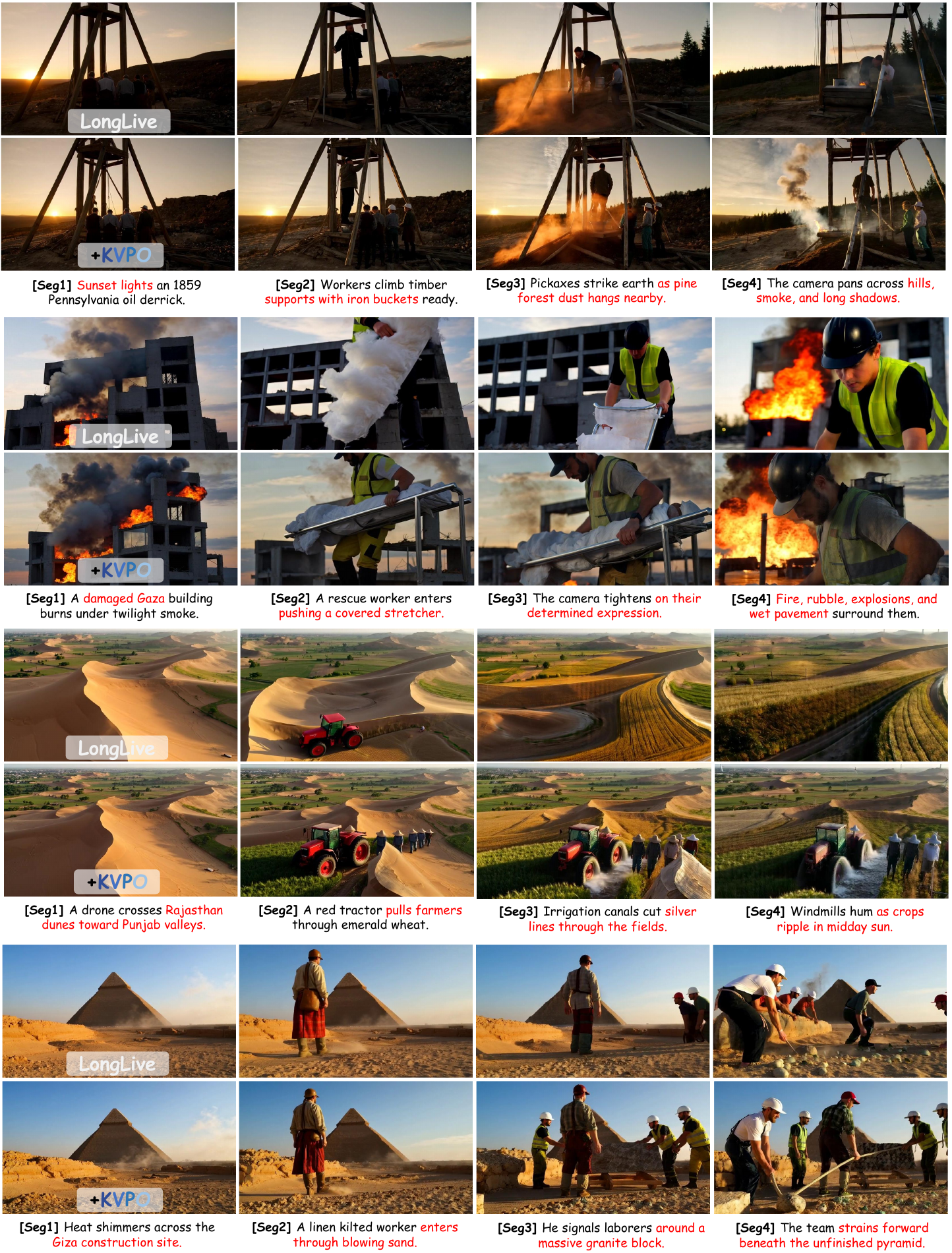}
    \caption{\textbf{Additional qualitative results on \textbf{LongLive}.}}
    \label{fig:app-longlive-4}
\end{figure*}

\begin{figure*}[p]
    \centering
    \includegraphics[width=\textwidth]{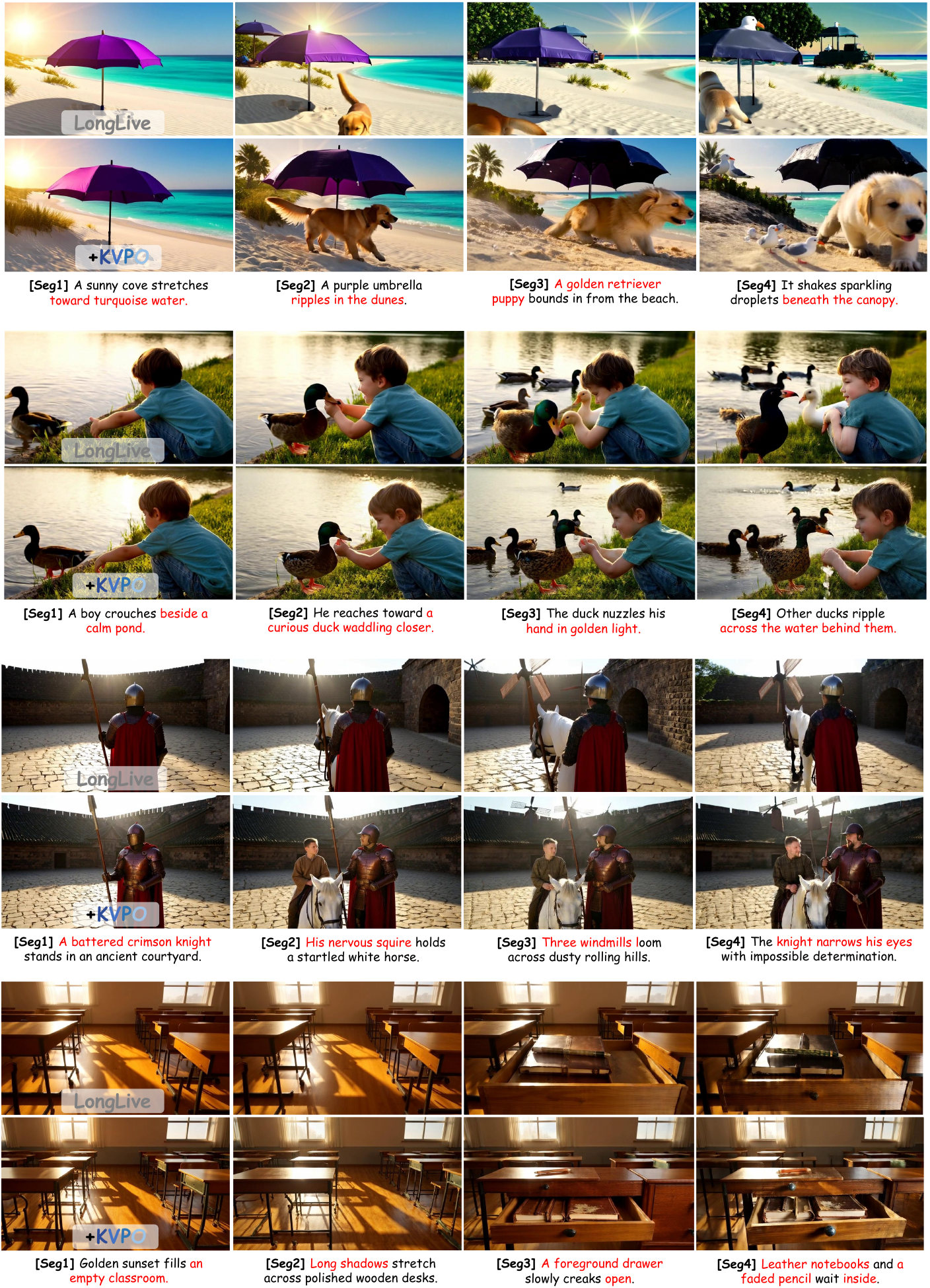}
    \caption{\textbf{Additional qualitative results on \textbf{LongLive}.}}
    \label{fig:app-longlive-5}
\end{figure*}

\begin{figure*}[p]
    \centering
    \includegraphics[width=\textwidth]{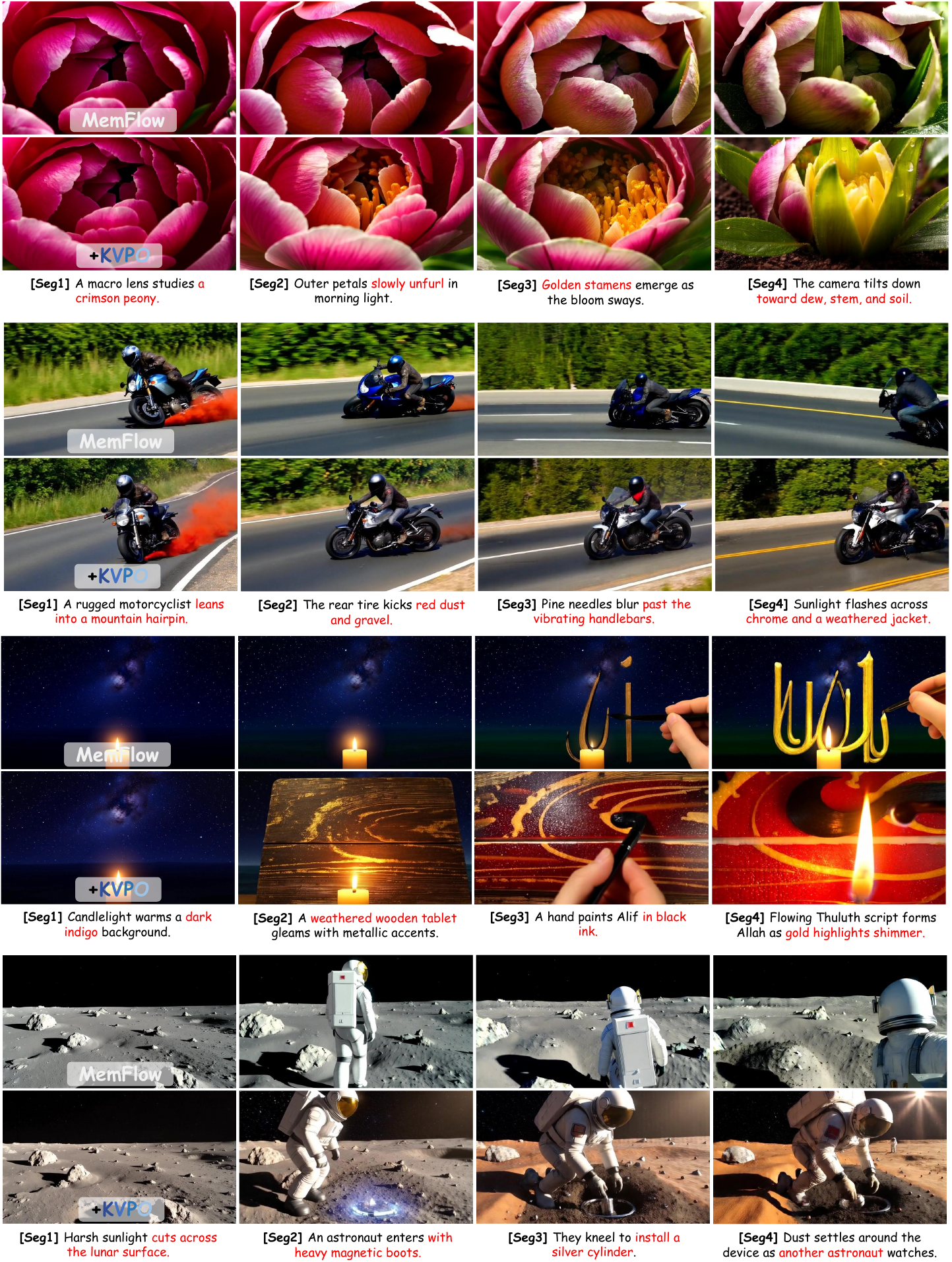}
    \caption{\textbf{Additional qualitative results on \textbf{MemFlow}.}}
    \label{fig:app-memflow-1}
\end{figure*}

\begin{figure*}[p]
    \centering
    \includegraphics[width=\textwidth]{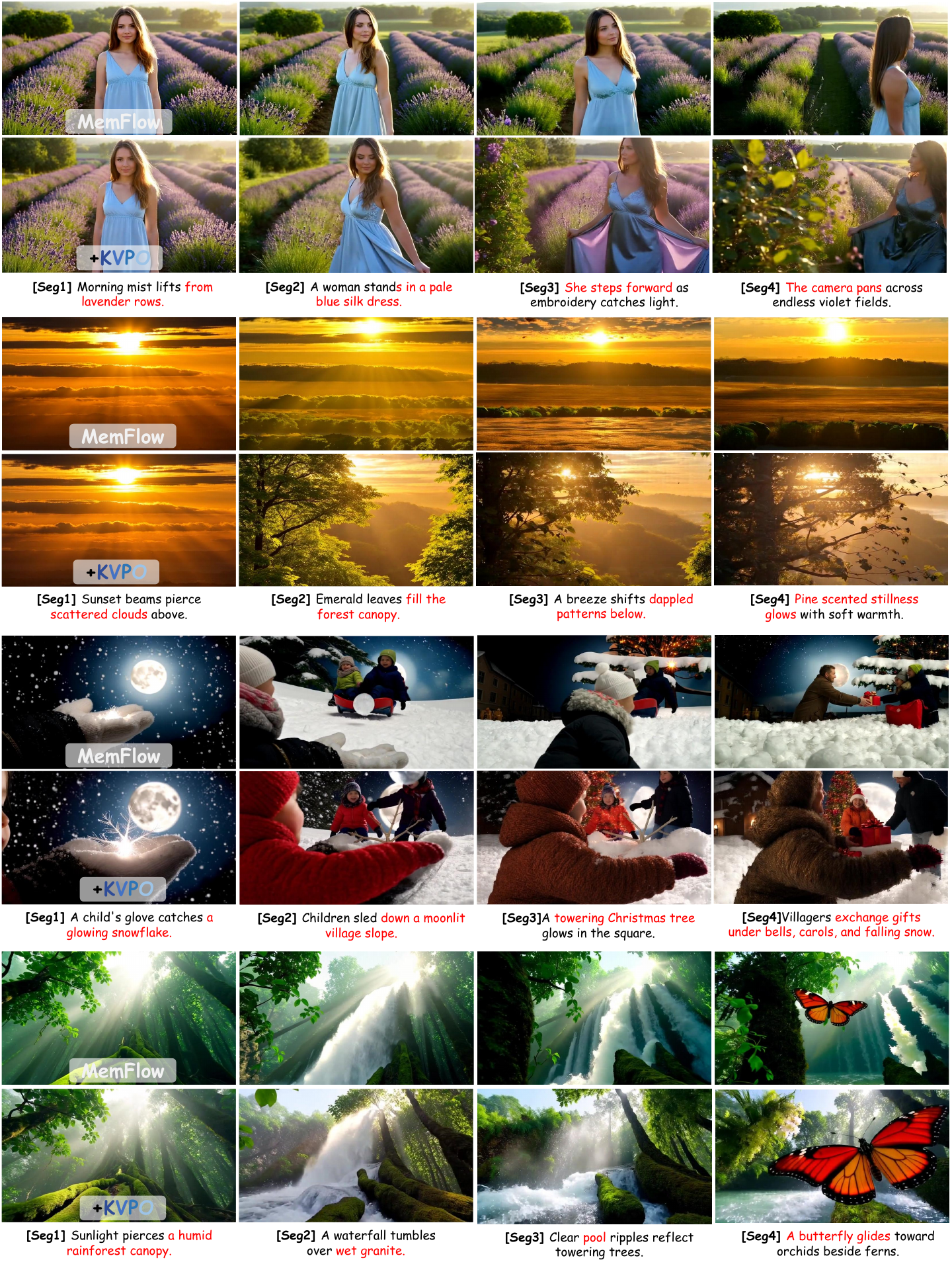}
    \caption{\textbf{Additional qualitative results on \textbf{MemFlow}.}}
    \label{fig:app-memflow-2}
\end{figure*}

\begin{figure*}[p]
    \centering
    \includegraphics[width=\textwidth]{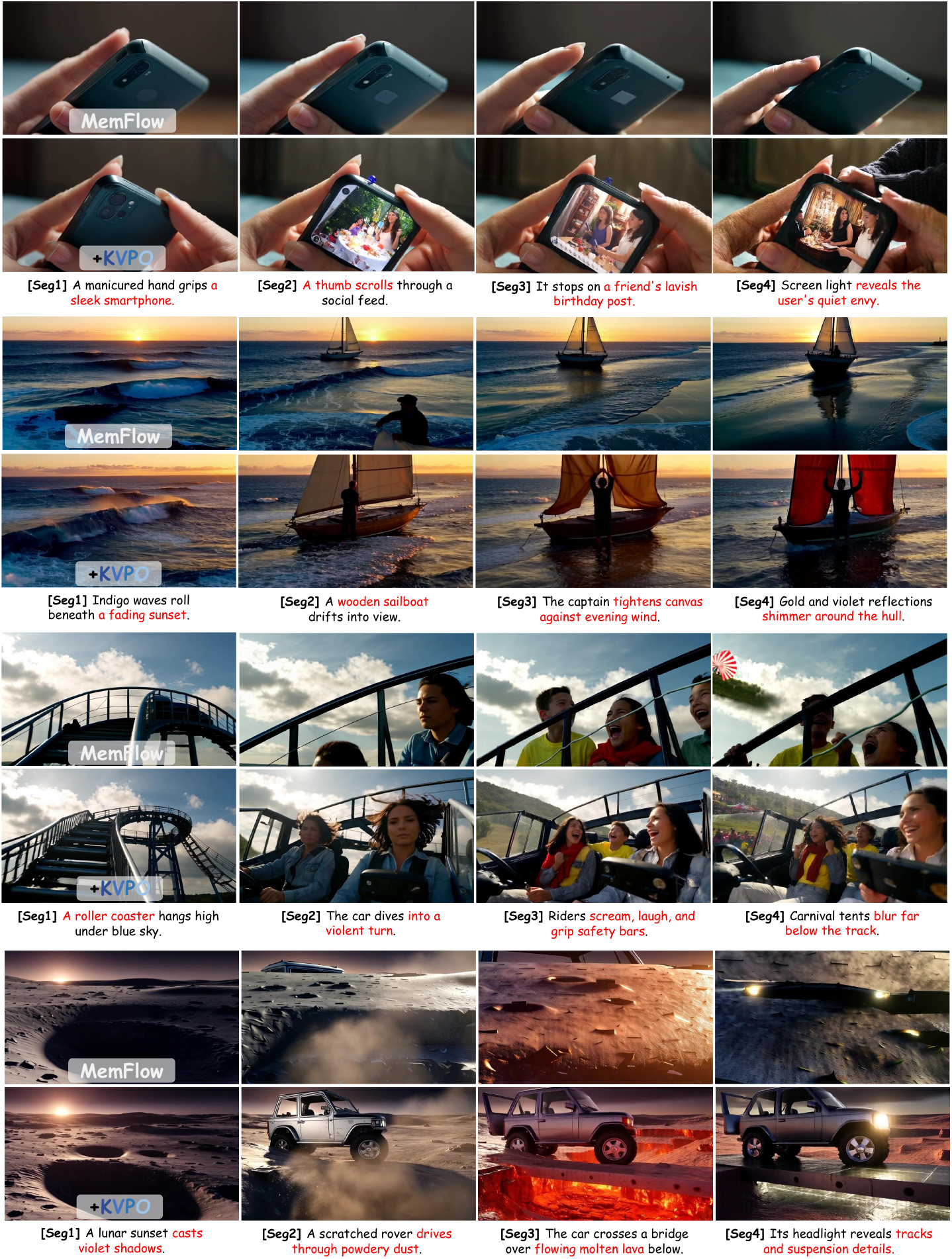}
    \caption{\textbf{Additional qualitative results on \textbf{MemFlow}.}}
    \label{fig:app-memflow-3}
\end{figure*}

\begin{figure*}[p]
    \centering
    \includegraphics[width=\textwidth]{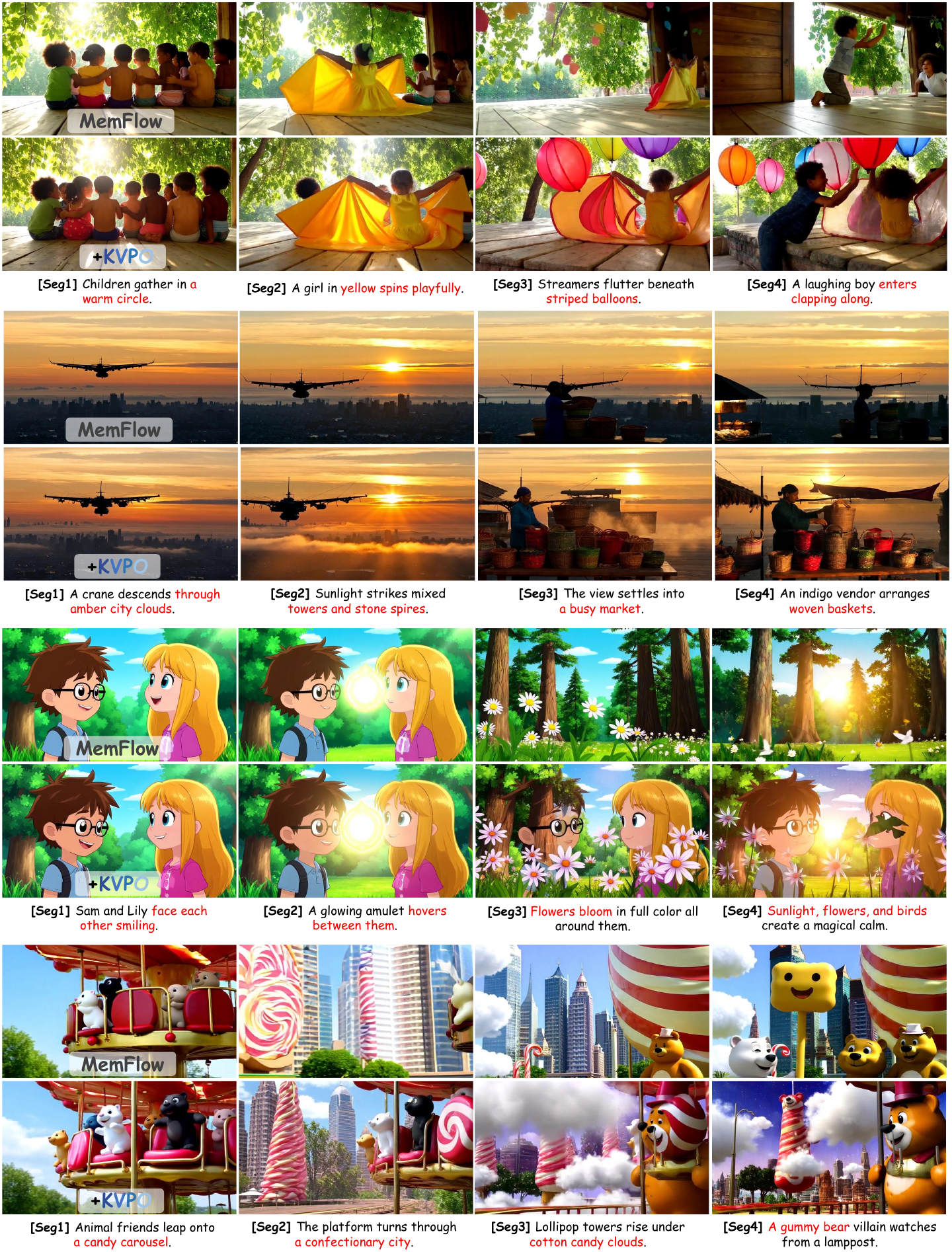}
    \caption{\textbf{Additional qualitative results on \textbf{MemFlow}.}}
    \label{fig:app-memflow-4}
\end{figure*}

\begin{figure*}[p]
    \centering
    \includegraphics[width=\textwidth]{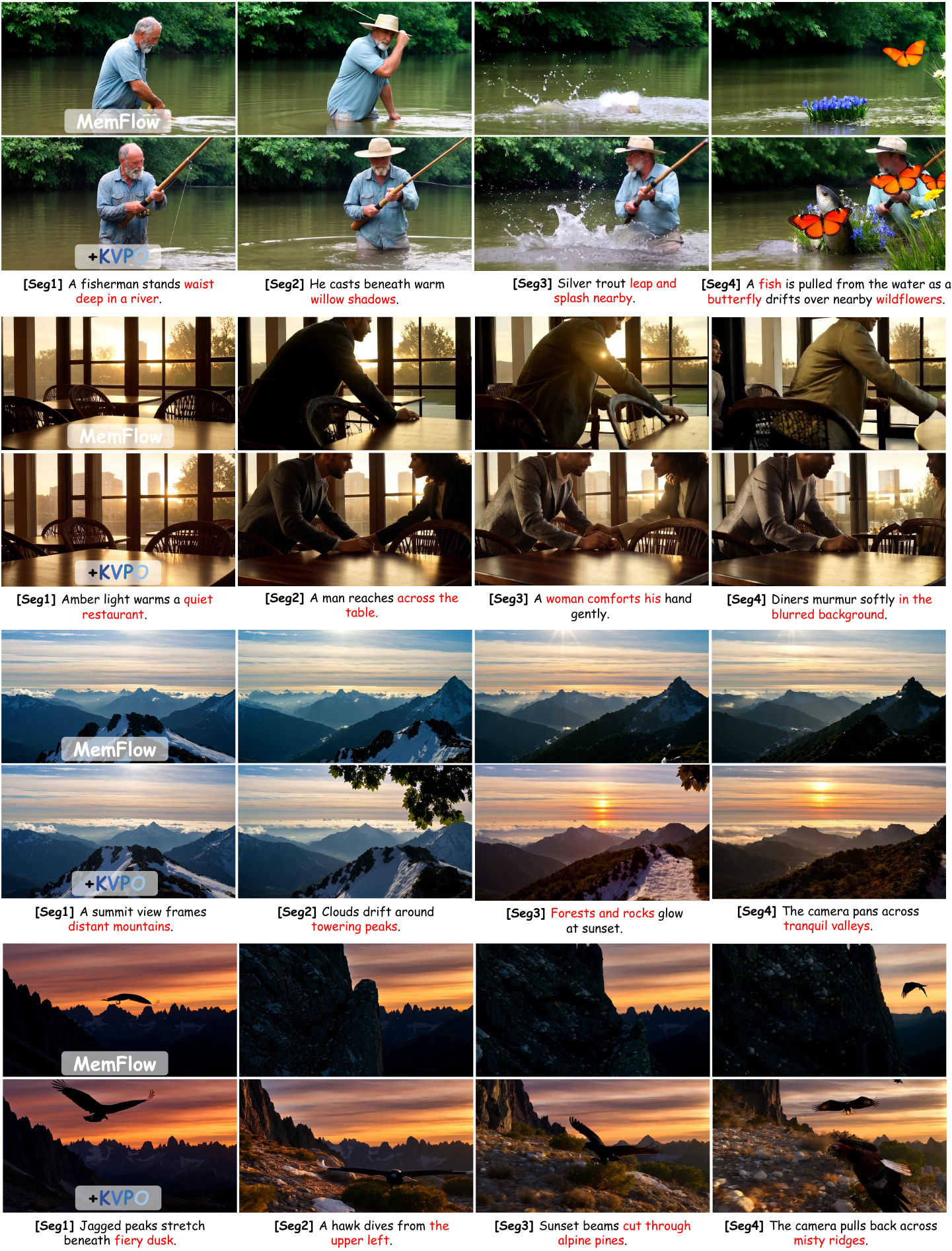}
    \caption{\textbf{Additional qualitative results on \textbf{MemFlow}.}}
    \label{fig:app-memflow-5}
\end{figure*}

\begin{figure*}[p]
    \centering
    \includegraphics[width=\textwidth]{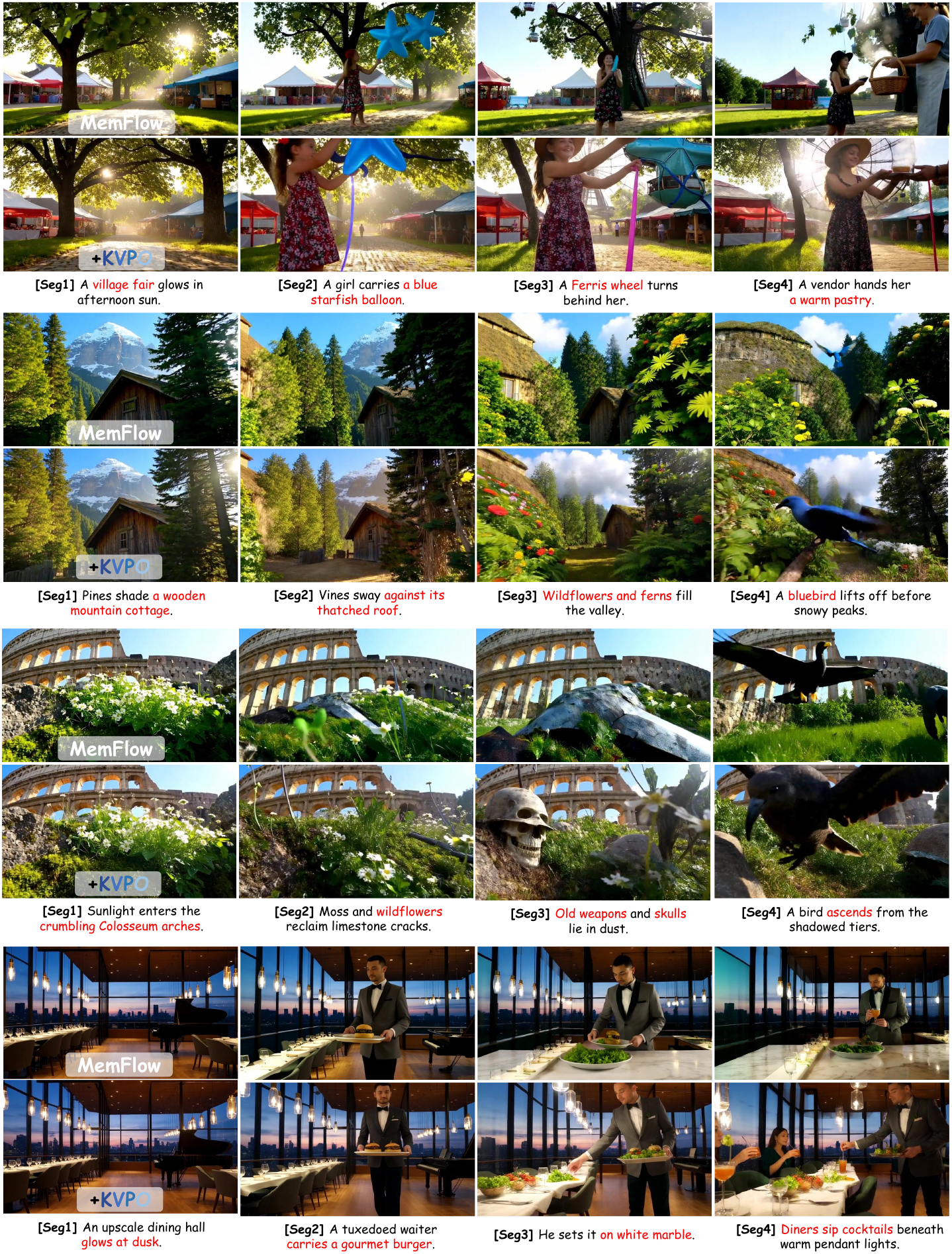}
    \caption{\textbf{Additional qualitative results on \textbf{MemFlow}.}}
    \label{fig:app-memflow-6}
\end{figure*}

\end{document}